\documentclass[lettersize,journal]{IEEEtran}
\usepackage{amsmath,amsfonts}
\usepackage{algorithmic}
\usepackage{algorithm}
\usepackage{array}
\usepackage[caption=false,font=normalsize,labelfont=sf,textfont=sf]{subfig}
\usepackage{textcomp}
\usepackage{stfloats}
\usepackage{url}
\usepackage{verbatim}
\usepackage{graphicx}
\usepackage{cite}
\usepackage{dsfont}
\usepackage{mathtools}
\usepackage{bm}
\usepackage{hyperref}

\begin{document}

\title{Uncertainty-Aware DRL for Autonomous Vehicle Crowd Navigation in Shared Space
}

\author{Mahsa Golchoubian$^{1\dag}$, Moojan Ghafurian$^{1}$, Kerstin Dautenhahn$^{2}$, Nasser Lashgarian Azad$^{1}$
\thanks{\dag Corresponding Author: Mahsa Golchoubian {\tt\small mahsa.golchoubian@uwaterloo.ca}}
\thanks{$^{1}$Department of Systems Design Engineering, University of Waterloo, Canada}
\thanks{$^{2}$Department of Electrical and Computer Engineering, University of Waterloo, Canada}%
}

\maketitle

\begin{abstract}

Safe, socially compliant, and efficient navigation of low-speed autonomous vehicles (AVs) in pedestrian-rich environments necessitates considering pedestrians' future positions and interactions with the vehicle and others. Despite the inevitable uncertainties associated with pedestrians' predicted trajectories due to their unobserved states (e.g., intent), existing deep reinforcement learning (DRL) algorithms for crowd navigation often neglect these uncertainties when using predicted trajectories to guide policy learning. This omission limits the usability of predictions when diverging from ground truth. This work introduces an integrated prediction and planning approach that incorporates the uncertainties of predicted pedestrian states in the training of a model-free DRL algorithm. A novel reward function encourages the AV to respect pedestrians' personal space, decrease speed during close approaches, and minimize the collision probability with their predicted paths. Unlike previous DRL methods, our model, designed for AV operation in crowded spaces, is trained in a novel simulation environment that reflects realistic pedestrian behaviour in a shared space with vehicles. Results show a 40\% decrease in collision rate and a 15\% increase in minimum distance to pedestrians compared to the state of the art model that does not account for prediction uncertainty. Additionally, the approach outperforms model predictive control methods that incorporate the same prediction uncertainties in terms of both performance and computational time, while producing trajectories closer to human drivers in similar scenarios.
\end{abstract}


\section{Introduction}

As more autonomous vehicles (AVs) and robots are being introduced into our everyday life, the importance of their proper navigation among pedestrians in crowded environments is getting more critical, leading to increasing research focused on human-aware or crowd navigation~\cite{kruse2013human, mavrogiannis2023core}.

The focus in this domain has primarily centred around the navigation of small mobile robots among pedestrians \cite{kruse2013human, mavrogiannis2023core}, with less attention being paid to the operation of low-speed AVs in analogous settings. In the future, AVs are expected to enter environments beyond conventional road structures, serving as mobility aids in places such as airport terminals, shopping malls, or shared urban spaces where the segregation between vehicles and pedestrians is minimal~\cite{predhumeau2021pedestrian}. This necessitates the development of navigation algorithms for AVs capable of handling diverse and complex human-vehicle interactions, aligning with the crowd navigation research area.

Earlier methods proposed for crowd navigation where based on reactive collision avoidance methods such as Optimal Reciprocal Collision Avoidance (ORCA) \cite{van2011reciprocal} or social force model~\cite{helbing1995social}. However, reactive methods, which rely on one-step interaction rule solely based on the current state of the agents are shortsighted and can lead to unnatural, and oscillatory trajectories~\cite{ferrer2013social, kretzschmar2016socially}.

\begin{figure}[t!]
  \centering
  \includegraphics[width=0.8\linewidth]{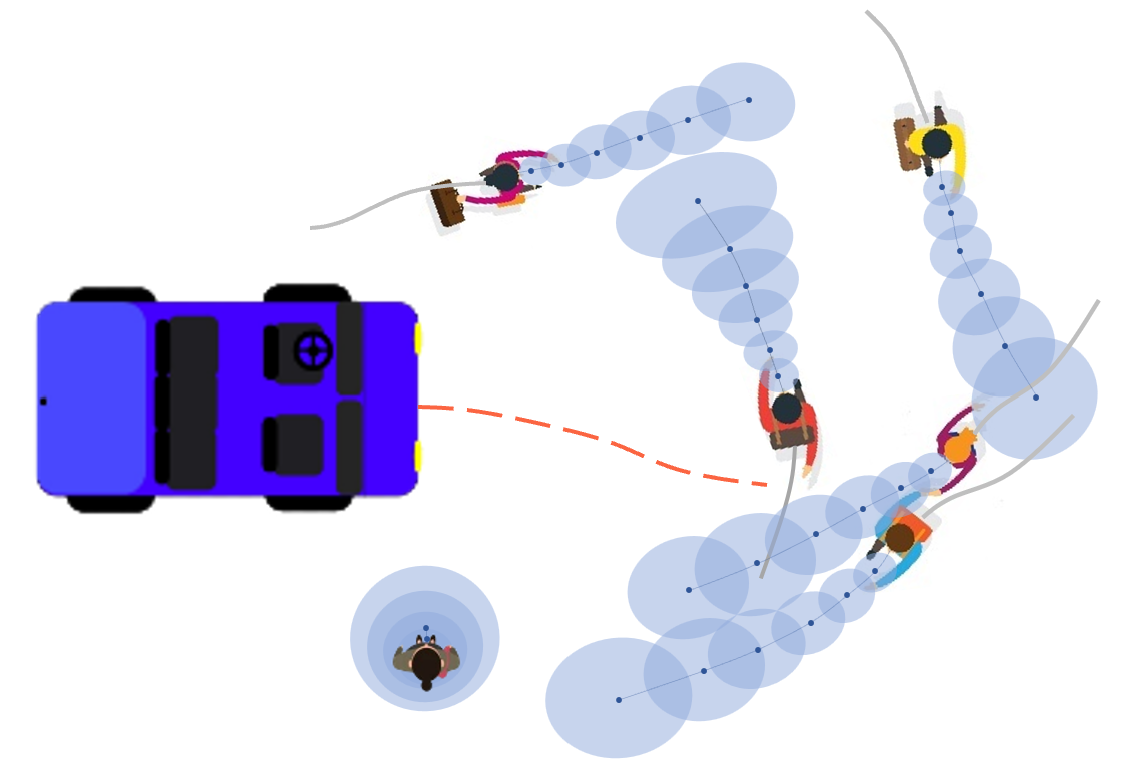}
  \caption{ We propose the use of predicted position and its associated covariance in the training of a deep reinforcement learning model with a novel reward function. Through this uncertainty-aware coupled prediction and planning approach, our model learns safe, foresighted, and smooth navigation behaviours among pedestrians in a shared space.}
  \label{fig:Overall}
\end{figure}

Proactive navigation algorithms, on the other hand, incorporate reasoning about the dynamic surroundings by employing prediction algorithms to generate more farsighted behaviours. Within this category, the subcategory of decoupled prediction and planning approaches predicts all agents' trajectories and attempt to plan the robot's path in the remaining unoccupied space~\cite{unhelkar2015human, du2011robot, bennewitz2005learning, aoude2013probabilistically}. However, decoupled methods often encounter the robot freezing problem in crowded settings, where the prediction renders most of the area unsafe to traverse~\cite{trautman2010unfreezing}. Overcoming this problem requires considering the interaction effect of agents on each other's trajectories and their collaborative path adjustment~\cite{trautman2010unfreezing}. As a result, there is another subcategory of coupled prediction and planning approaches that jointly plan feasible trajectories for all agents~\cite{kretzschmar2016socially,trautman2015robot, kuderer2012feature}, considering the adjustment of the preferred trajectory distribution during the interaction \cite{Sun2021}. However, implementing this joint planning approach in crowded settings, coupled with the need for frequent updates to compensate for deviations in other agents' paths from the jointly planned trajectory, imposes a computational time burden, making them challenging for real-time implementation~\cite{chen2017decentralized}.

Hence, in recent years, a third category of learning algorithms has gained significant attention~\cite{chen2017decentralized,chen2019crowd,liu2023intention}. This category efficiently manages computational time by conducting an offline training phase, followed by the utilization of the trained model in a rapid inference phase~\cite{chen2017decentralized}. Learning-based navigation methods within this category can grasp the in-between interaction effects for socially-aware crowd navigation by interacting with a simulated environment populated by pedestrians using Deep Reinforcement Learning (DRL)~\cite{chen2017decentralized,chen2019crowd,liu2023intention,xie2023drl}.

However, there are several limitations to the existing DRL models designed for crowd navigation: (i) They often rely solely on the current state of all agents, neglecting the benefits of incorporating predicted human trajectories or, when considering predicted trajectories, they ignore the uncertainties of pedestrians' behaviour, resulting in a potentially misleading reliance on prediction; (ii) Current reward functions have limitations in incorporating pedestrian comfort, particularly regarding the distance the robot should keep from pedestrians and the impact of robot's speed at close passes on pedestrian discomfort. (iii) Simulation environments used for training focus on robots close in size to pedestrians and make strong assumptions about pedestrian behaviours based on the ORCA model, which may not reflect real-world behaviours. 

To overcome these limitations, we introduce an uncertainty-aware DRL-based coupled prediction and planning model designed for navigating a low-speed AV within a shared space with pedestrians (see Fig.~\ref{fig:Overall}). To address the first limitation of not accounting for prediction or their uncertainties, we enhance the model by incorporating not only the predicted future trajectories of pedestrians but also accounting for the uncertainties associated with these predictions in the formulation of a model-free DRL framework. To tackle the second limitation regarding pedestrian comfort, we propose a novel reward function component for the DRL model. This addition aims to encourage the AV to learn improved social behaviour. Finally, the last limitation concerning simulation environments, is addressed by training our model in a simulation environment derived from realistic pedestrian-vehicle interactions in a shared space. This simulation environment is constructed using data from the Hamburg Bergedorf Station (HBS) dataset \cite{pascucci2017discrete,pascucci2021dataset}, to ensure a more realistic representation of pedestrian behaviours in shared spaces.

In summary, our contributions are listed below:

\begin{itemize}

    \item Integration of model-free DRL crowd navigation with uncertainty-aware pedestrian trajectory prediction, using a unique reward function that guides the AV to avoid intruding pedestrians' current positions and minimize collision probabilities with their predicted future paths.
    
    \item Introduction of a novel reward function that prioritizes pedestrians' comfort by encouraging the AV to reduce speed during close approaches
    
    \item Addressing AV crowd navigation challenges by training in a novel simulation environment that reflects real pedestrian behaviour in shared spaces with vehicles.
    
    \item Introduction of a novel uncertainty-aware loss function for training the data-driven prediction model, aimed at improving the accuracy of forecasting the covariance of future positions within the predicted distribution.

\end{itemize}

\section{Related Works}

\subsection{Crowd navigation using Deep Reinforcement Learning}

Deep Reinforcement Learning (DRL) algorithms have been proposed as an effective means for understanding the complex interaction effects among agents in robot crowd navigation with reasonable computational time during execution~\cite{chen2017decentralized}. The initial methods were based on value function learning (V-learning), requiring a known state transition function for policy extraction~\cite{chen2017decentralized, chen2017socially, chen2019crowd}. Consequently, assumptions regarding pedestrian behaviour were introduced into these models, such as adhering to a constant velocity model \cite{chen2017decentralized} or conforming to the ORCA model~\cite{chen2019crowd}.

Subsequent models have evolved to relax these assumptions. Some approaches directly learn both a policy function and a value function within an actor-critic model-free DRL architecture~\cite{everett2018motion, liu2021decentralized}. Others aim to enhance the models by incorporating the learning of the state transition function using prediction models, gradually moving towards a more model-based DRL approach~\cite{chen2020relational, matsumoto2022mobile, li2020socially}. In the latter scenario the search for optimal actions based on the value function and the prediction model is restricted to a small discrete set of actions. Additionally, to manage computational time, the consideration of future steps for maximizing returns in action selection typically remains limited to one or two steps~\cite{chen2020relational,matsumoto2022mobile}.

However, pedestrian trajectory predictions can also contribute to informing model-free DRL models. While certain methods in this category rely solely on one-step prediction \cite{sathyamoorthy2020densecavoid}, Liu et al. have advanced the approach by employing predictions over a future horizon to guide the robot away from pedestrians' intended future positions~\cite{liu2023intention}. Notably, a gap in the existing literature is the lack of consideration for the impact of uncertainties in these predictions on the learned policy. Addressing prediction uncertainties is especially valuable for scenarios involving pedestrians suddenly entering the robot or AV's field of view without many past observable states of motion or for stationary pedestrians who might unexpectedly initiate movement. These instances represent cases where data-driven predictions are prone to large errors. Therefore, in our proposed method, we account for these uncertainties to enhance the risk awareness of the learned DRL policy.

\subsection{Uncertainty-aware motion planning}

The incorporation of uncertainties from different sources has been studied across various modules of an autonomous navigation pipeline, spanning from perception \cite{xu2023model} to control \cite{yang2023uncertainties}. Within the decision-making module, the uncertainties related to predicted trajectories of other vehicles during AV's road operation, are considered in \cite{hubmann2018automated,xiong2023integrated} by planning in the belief space using a partially observable Markov decision process (POMDP) in scenarios involving navigation through unsignalized intersections and lane changes where interaction with other agents are crucial. 
In addressing the uncertainty regarding the behaviour of other agents at an unsignalized intersection, a multi-modal interaction-aware prediction model is proposed in \cite{trentin2023multi} by combining Markov Chains and Dynamic Bayesian Networks (DBNs). Their multi-modal prediction takes into account the various intentions of nearby agents, which are inferred through the utilization of a DBN.

Others have integrated uncertainties of the predicted states of surrounding vehicles into the constraint formulation of a model predictive control (MPC) method~\cite{zhou2023interaction,brudigam2021stochastic,akhtyamov2023social}. 

Among the deep reinforcement learning algorithms used for decision making and motion planning, Wu et al, have incorporated the  uncertainties associated with their environment transition model into a model-based DRL framework~\cite{wu2022uncertainty}.

However, the incorporation of diverse predictions for pedestrians' trajectories and the associated uncertainties is limited in existing crowd navigation methods. Li et al. introduced the utilization of multiple plausible predictions generated by a multimodal prediction algorithm within the V-learning method~\cite{li2020socially}. In their approach, the value function is evaluated for all possible future predictions, and the best action is selected from a discrete set. Despite considering multiple predictions, each anticipated trajectory consists of a deterministic future position for pedestrians, without assigning a probability to each of the multiple predictions. Furthermore, in their method, the use of different predictions is limited to the value function calculation and does not extend to the state or reward function formulation~\cite{li2020socially}.

In contrast, our proposed method leverages the use of predicted states and their uncertainties in both state construction and the reward function. We consider a continuous action space within a model-free policy-based deep reinforcement learning (DRL) framework, capable of producing smoother trajectories compared to cases involving discrete action considerations. Our uncertainties are represented by the covariance matrix of the predicted bivariate Gaussian distribution of pedestrians' future positions, generated by a data-driven trajectory predictor trained on real trajectory data.

\begin{figure}[t!]
  \centering
  \includegraphics[width=1.0\linewidth]{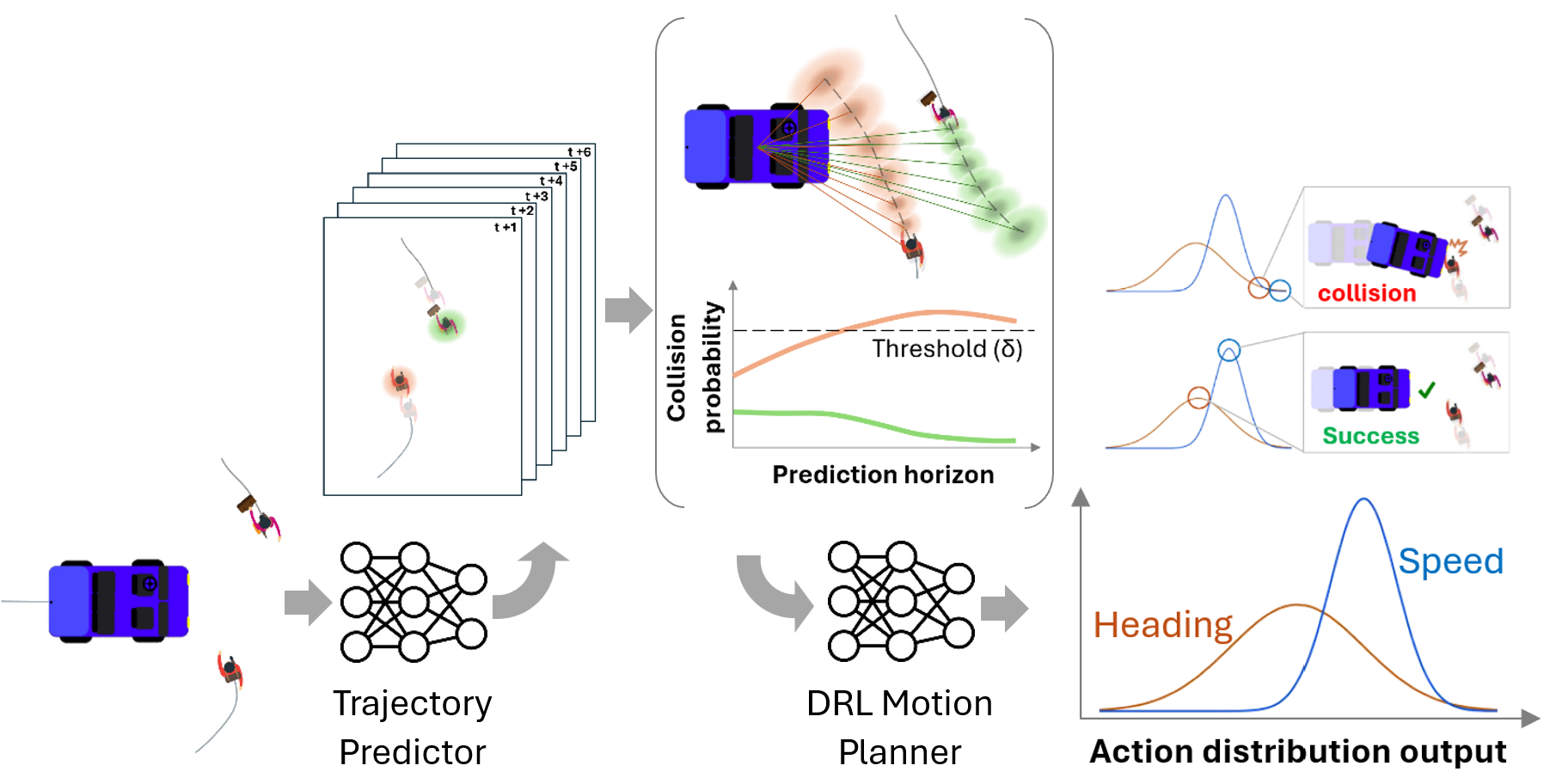}
  \caption{Overview of the proposed framework. At each time step, current observations are input to the predictor, generating pedestrians' predicted positions and associated uncertainties over a horizon. These predictions inform collision probability calculations between the AV and each pedestrian's predicted path, contributing to the  reward function of the DRL module. Utilizing this predictive data, the DRL motion planner is trained to generate a probability distribution of optimal actions, mitigating collision risk with both current and intended pedestrian paths.}
  \label{fig:illustration}
\end{figure}

\section{Approach}

\subsection{Problem Formulation}

The problem of navigating from a start point to a goal position among a crowd of pedestrians can be formulated as a sequential decision making process using the Markov Decision Process (MDP) notation defined through the tuple $<S,A,P,R,\gamma>$. Let $w_t$ be the AV's current state and $s_t^i$ be the $i$-th pedestrian's current state. The state of the AV consists of its current position ($p^{AV}_t$), velocity, heading angle, size (specified as radius $r^{AV}$), preferred velocity and goal position ($w: [p_x, p_y, v_x, v_y, \theta, r, v_{pref}, g_x, g_y]$). We define pedestrian i's current state as its current position $s^{i}_t$: ($p_x^{i}, p_y^{i})$. Considering the pedestrians position history over the last H time steps ($s^{i}_{t-H:t}$), we also predict the pedestrians future position and its covariance over the next K time steps as $\hat{s}^{i}_{t+1:t+K}$ and $\hat{\Sigma}_{t+1:t+K}^{i}$ with the ground truth (GT) predictions being denoted as $s^{i}_{t+1:t+K}$. Therefore, we define the state of the MDP, $s_t \in S$ as the joint state of the AV and the \textit{n} pedestrians states including their current and predicted state and covariance: $s^{jn}_t: (w_t, s^{1}_t, \hat{s}^{1}_{t+1:t+K}, \hat{\Sigma}_{t+1:t+K}^{1}, ..., s^{n}_t, \hat{s}^{n}_{t+1:t+K},$ $ \hat{\Sigma}_{t+1:t+K}^{n})$. Pedestrians modelled as circles have a radius denoted by $r^{ped}$. Considering a unicycle model for the AV, we define the action $a_t \in A$ as the speed and heading change $a_t = [v_t, \Delta \theta]$. Taking the action at each time step the AV receives a reward $r_t \in R$ and the AV and the whole environment moves to the next step according to an unknown state transition probability $P(s^{jn}_{t+1} | s^{jn}_t , a_t)$. The objective is to find the optimal policy $\pi(a_t |s^{jn}_t)$ that maps the joint state to an action while maximizing the cumulative discounted reward ($\sum_{k=0}^{\infty} \gamma^k r_{t+k}$) with a discount factor of $\gamma$. The navigation process will continue until (a) the AV reaches the goal, (b) collides with any pedestrians or (c) the simulation exceed the allowed time given to the AV to reach the goal. 
Figure \ref{fig:illustration} depicts an overview of the proposed framework, with detailed descriptions of each module provided in subsequent sections.

\begin{figure*}[t!]
  \centering
  \includegraphics[width=0.8\linewidth]{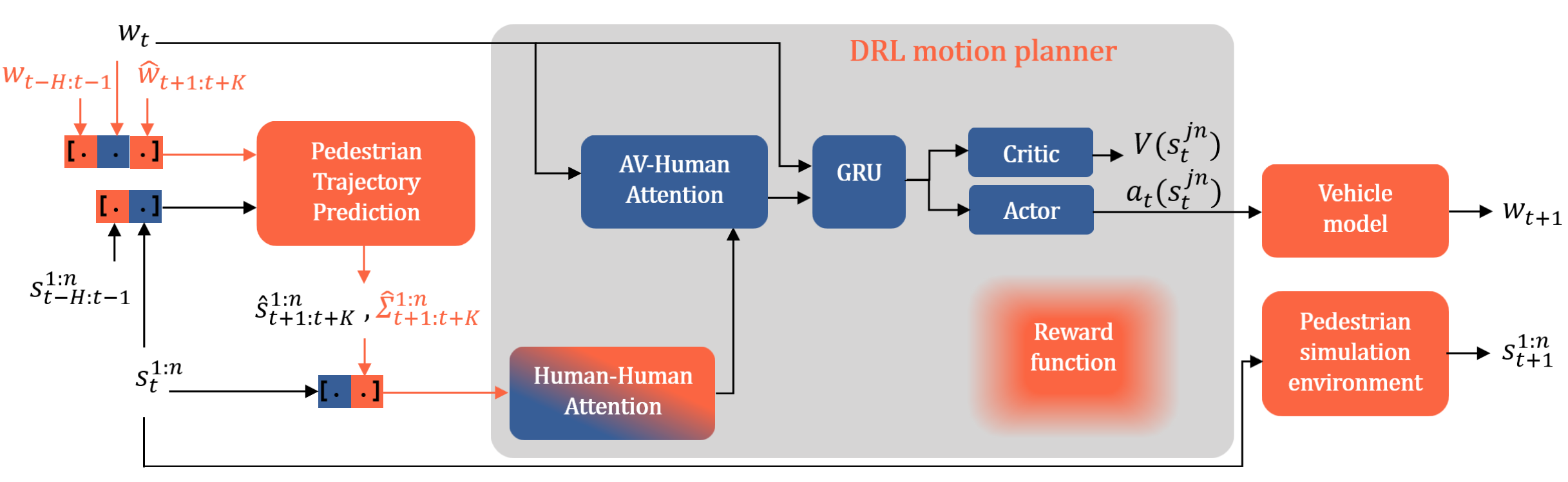}
  \caption{Our integrated prediction and planning framework. The DRL network architecture is based on the model introduced in \cite{liu2023intention}, with novel additions and modifications highlighted in orange. The inclusion of prediction uncertainties (covariance) is a key feature, integrated both as part of the observation state and in the formulation of our novel reward function. This covariance is obtained from the predictor module, which addresses pedestrian-vehicle interaction by incorporating the AV's state and an estimate of its future path as input. In our model, prediction covariance, alongside the states themselves, contributes to the human-human attention module.}
  \label{fig:model}
\end{figure*}

\subsection{Trajectory Predictor}

As explained, our joint state also includes the predicted position and covariance of nearby pedestrians over a prediction horizon. This information is provided by the pedestrian trajectory predictor, which is integrated with the DRL network. Recognizing the significance of nearby agents' influence on pedestrians' future trajectories \cite{golchoubian2023pedestrian}, the predictor takes into account the interactions between pedestrians and vehicles in producing its prediction. By leveraging the history of pedestrians' trajectories for H steps, the predictor generates the mean vector ($\overrightarrow{\mu}$) and the covariance matrix ($\hat{\Sigma}$) of the pedestrians' predicted positions over the next K steps as a bivariate Gaussian distribution, from which the predicted positions are sampled (Eq. \ref{eq:pred}).

\begin{equation} \label{eq:pred} 
\begin{split}
    \hat{s}^{i}_{t+1:t+K} = \mathcal{N}(\overrightarrow{\mu}_{t+1:t+K}^{i}, \hat{\Sigma}_{t+1:t+K}^{i}) = \\
    Predictor(s^{1:n}_{t-H:t}, w_{t-H:t})
\end{split}
\end{equation}

We employed our previously developed data-driven pedestrian trajectory predictor, Polar Collision Grid (PCG) \cite{golchoubian2023polar}, trained on HBS dataset, which contains real-world trajectory data from pedestrian-vehicle interactive scenarios in a shared space \cite{pascucci2017discrete}. The predictor's architecture is specifically designed to capture in-between interaction effects. The pre-trained PCG predictor is used in its inference mode to generate the predicted trajectories. It takes both pedestrian and AV's trajectories as inputs, allowing it to account for the impact of AV's presence on pedestrians' future trajectories. This approach contrasts with previous works that often assume that the robot is invisible to humans when generating predictions for use in the DRL formulation \cite{liu2023intention, katyal2020intent, li2020socially}.

Given that the covariance of the predicted trajectory (as a measure of prediction uncertainty) is also used in the observation state of the DRL formulation, good prediction of the covariance alongside the actual $x$ and $y$ position of the data sampled from it is also very important. This has encouraged us to use another trained model of the original Polar Collision Grid predictor with a revised loss function. In this modified version, in addition to the common negative log likelihood loss (NLL), we also add an uncertainty loss aimed at improving covariance prediction without compromising accuracy in mean prediction. This adjustment is essential as a model trained solely with the NLL loss is prone to providing overconfidence predictions, where the output distribution aligns precisely with the ground truth predicted position, with a sharp distribution similar to a Dirac delta distribution \cite{ivanovic2022propagating}. This implies a low standard deviation for the predicted distribution or, in other words, very low uncertainty.

Hence, employing the combined loss in Eq.~(\ref{eq:loss}) directs the training process not only to maximize the log probability of the ground truth position but also to consider the distance between the ground truth point and the predicted distribution. This encourages predictions to have a higher standard deviation for the same predicted mean, aiming to encompass the ground truth.

\begin{equation}\label{eq:loss}
\begin{split}
     Loss = - \sum_{i,t} log(P(s_{t+1}^{i} | \overrightarrow{\mu}_{t+1}^{i}, \hat{\Sigma}_{t+1}^{i})) + \\
    W * \sum_{i,t} d^{MD}(s_{t+1}^{i}, \overrightarrow{\mu}_{t+1}^{i}, \hat{\Sigma}_{t+1}^{i})
\end{split}
\end{equation}

In this equation, $d^{MD}$ represents the Mahalanobis distance calculated between the ground truth (GT) position of the pedestrians and the predicted normal distribution. The uncertainty loss function is introduced as an addition to the original negative log likelihood (NLL) loss function, with a weight denoted as $W$. The outcomes of the model trained with this combined loss function, in comparison to the original model trained solely with NLL, are presented in section~\ref{res-pred}.

\subsection{DRL Network Architecture}

A model-free DRL network is used as the AV's decision-making algorithm. We built upon the DRL network architecture proposed in \cite{liu2023intention}, which represents interactions among agents using edges in a spatio-temporal graph structure. This architecture captures both the direct impact of pedestrian-AV interactions and the indirect influence of pedestrian-pedestrian interactions on AV's decision-making (Fig. \ref{fig:model}). These effects are modelled and aggregated through attention mechanisms. The temporal correlation of the graphs at consecutive time steps, referred to as temporal edges, is captured using a gated recurrent unit (GRU). The GRU outputs features to both a value network and a policy network, both of which are trained together using Proximal Policy Optimization (PPO). To model the limited sensor range of the AV, all pedestrians within a specified range are considered to construct the spatial edges.
 
However, our model differs from the model presented in \cite{liu2023intention} in four key aspects, as depicted in Fig. \ref{fig:model}. Firstly, our predictor not only generates the covariance for the predicted position of pedestrians but also integrates it as part of the joint state input for the DRL policy generator, effecting the attention module that captures agent interactions. The predicted covariance also contributes to formulating the reward function during training. Secondly, unlike the integrated prediction and planning model proposed in \cite{liu2023intention}, our pedestrian trajectory predictor incorporates the AV's state as input, capturing the influence of the AV on the predicted pedestrian trajectory. In line with this approach, we use the AV's most recent action to project its trajectory with a constant velocity model, providing an estimated trajectory for the AV over the next K steps ($\hat{w}_{t+1:t+K}$). 
This projection serves as a simple estimation of how pedestrians are likely to perceive the AV's trajectory in the near future. Subsequently, this estimated future trajectory along with the current AV's observed trajectory, is inputted into the pedestrian trajectory predictor module to model pedestrian-AV interactions within the predictor. It is important to note that the constant velocity projection of the AV's trajectory is solely utilized within the pedestrian trajectory prediction module for calculating AV-pedestrian interaction features. Based on these predictions, the DRL module outputs only the best action for the AV at the current time step. Once this single action is executed, and the environment transitions to the next time step, the entire process repeats with the newly available observations.

Thirdly, we adopt a unicycle kinematic model, which more accurately represents the non-holonomic kinematics of vehicles, as opposed to the holonomic kinematic model utilized in \cite{liu2023intention} for training the DRL network. Finally, our model is trained within our own developed novel simulation environment with pedestrian behaviour based on realistic human trajectories in a space shared with vehicles rather than assuming pedestrians comply with the ORCA model.

\subsection{Reward Function}

In the design of the reward function for the DRL network, the AV is given a high reward when it reaches the goal point within a predefined acceptable radius outlined in the goal set $S_{goal}$. Conversely, the system penalizes the AV not only for collisions with pedestrians but also for intruding into their Personal Space (PS), referred to as ``danger cases".

In all other instances, the AV is rewarded for its progress towards the goal ($r_{progress}$), penalized for large actions ($r_{action}$), and for coinciding with pedestrians' future predicted positions ($r_{pred}$). The reward function is outlined in Eq.~(\ref{eq:R}), where $d_{min}$ represents the minimum distance between the AV and all pedestrians. This is computed by subtracting the sum of the agents' radii from the Euclidean $L^2$-norm of the difference between their positions, where $\lVert . \rVert$ represents the Euclidean norm ( $d_{min} = \min \limits_{i} (\lVert p^{AV} - s^{i} \rVert - r^{AV} - r^{ped})$.

\begin{equation}\label{eq:R}
    r(s_t^{jn}, a_t) =
    \begin{cases}
      r_g & \text{if $s_t^{AV} \in S_{goal}$}\\
      r_c & \text{if $d_{min} < 0$}\\
      r_{danger} & \text{if $d_{min}< PS$} \\
      r_{progress}+r_{pred}+r_{action} & \text{otherwise}
    \end{cases}  
\end{equation}

To model pedestrians' discomfort, we explore two versions of danger penalties. In the first version, we adopt a widely used penalty formulation from the literature. Here, the AV incurs a linearly increasing penalty as its proximity to a pedestrian falls below a specified threshold—the closer the AV gets, the higher the penalty \cite{liu2023intention,liu2021decentralized,chen2019crowd}. However, this approach falls short in addressing the influence of the AV's speed on pedestrians' comfort zones \cite{camara2021space,chugo2013dynamic}, neglecting the individuals' tendency to maintain a larger distance from a vehicle travelling at higher speeds.

To address this limitation, we incorporate the AV's speed as a crucial factor in the second version of the danger penalty, referred to as speed-dependent (speed depend.). We introduce the AV's speed as a determinant in calculating the penalty, as outlined in Eq. \ref{eq:R_danger}. This modification is important particularly when modelling interactions in a human-vehicle mixed environment, recognizing the elevated risk linked to vehicles compared to smaller robots.

\begin{equation}\label{eq:R_danger}
    r_{danger} =
    \begin{cases}
      r_c (1 +\frac{v_{AV}}{v_{max}}) (\frac{PS - d_{min}}{PS}) & \text{if speed depend.}\\
      r_c (\frac{PS - d_{min}}{PS}) & \text{otherwise}\\
      
    \end{cases}  
\end{equation}

The progress reward ($r_{progress} = (-d_{goal}^t + d_{goal}^{t-1})$), is aimed at guiding the AV towards the goal position, with $d_{goal}$ representing the distance of the AV from the goal position. The action reward ($r_{action}$) is to penalize the AV for large heading changes as well as backward motion for smoother trajectories.

The prediction reward ($r_{pred}$) involves penalizing the AV if its position coincides with the pedestrians' intended future position, as per their predicted trajectory. This aspect of the reward function was initially introduced in \cite{liu2023intention}. However, in contrast to their deterministic prediction, our approach incorporates the uncertainties associated with predicted trajectories, embedded in the covariance matrix of the predicted distribution, when formulating the prediction reward. 
Consequently, we impose a penalty on the AV when the collision probability between its current state and the predicted position distribution of pedestrian \textit{i} exceeds a threshold $\delta$, based on Eq.~(\ref{eq:P_Col}) \cite{du2011probabilistic}. In this equation $V_s$ is the volume of the sphere of radius $r^{ped} + r^{AV} + PS$. We indicate the occurrence of this condition using the indicator $\mathds{1}_{t+k}^i$ (Eq. (\ref{eq:indicator})), for all the pedestrians' predicted state distribution over the next K time steps, with a decreased penalty for collisions in steps further into the future as per Eq.~(\ref{eq:R_pred}).

\begin{equation}\label{eq:P_Col}
\begin{aligned}
    & P_{t+k}^{i}(collision) \approx \\ 
    & V_s \frac{1}{\sqrt{det(2\pi\hat{\Sigma}_{t+k}^{i})}} \exp(-\frac{1} {2}d^{MD}(p_t^{AV}, \hat{s}_{t+k}^{i}, \hat{\Sigma}_{t+k}^{i})^2)
\end{aligned}
\end{equation}

\begin{equation}\label{eq:indicator}
    \mathds{1}_{t+k}^{i} = \begin{cases} 
    1 & \text{if $P_{t+k}^{i}(collision) > \delta$} \\
    0 & \text{otherwise} 
    \end{cases}
\end{equation}

\begin{equation}\label{eq:R_pred}
\begin{split}
    r_{pred}^{i}(s_t^{jn}) &= \min_{k=1,...,K}( \mathds{1}_{t+k}^{i} \frac{r_c}{2^k})\\
    r_{pred}(s_t^{jn}) &= \min_{i=1,...,n} r_{pred}^{i} (s_t^{jn})
\end{split}
\end{equation}

\subsection{MPC formulation}

Due to the higher safety criticality of a low-speed autonomous vehicle in a crowded environment, linked to their larger size and speed compared to small mobile robots, we conduct a comparison between our DRL model and methods from optimal control, where collision avoidance can be formulated as a constraint. In this context, we adopt the Model Predictive Control (MPC) method as a baseline, ensuring a fair comparison as it can be integrated with the pedestrian trajectory predictor in a manner similar to our DRL model. In the tests of the MPC navigation algorithm, the entire DRL motion planner block depicted in Fig. \ref{fig:model} will be substituted with the MPC-based motion planner.

We adopt the MPC formulation used for crowd navigation in \cite{akhtyamov2023social}, where the optimization problem is defined as follows:

\begin{equation}\label{eq:MPC}
\begin{aligned}
    & \min_{p_{1:K}, u_{0:K-1}} J_K(p_K^{AV}) + \sum_{k=0}^{K-1} J_k(a_k, p_k^{AV}) \\
\end{aligned}
\end{equation}

The cost in this formulation consist of a stage cost and a terminal cost, as defined in Eq. (\ref{eq:stage}) and Eq. (\ref{eq:terminal}) respectively.

\begin{equation}\label{eq:stage}
    J_k(a_k, p_k^{AV}) = J_k^a(a_k) + J_k^p(p_k^{AV}) + J_k^{ED}(p_k^{AV},s_k^{1:n})
\end{equation}

The stage cost consists of three components: a control input cost defined in Eq. (\ref{eq:J_u}), a normalized target distance cost outlined in Eq. (\ref{eq:J_p}), and a pedestrian closeness cost specified in Eq. (\ref{eq:peds}). The control input cost aims to minimize the usage of control signals. The normalized target distance cost serves a similar purpose as the progress-to-goal component in the reward function of the DRL formulation, while the pedestrian closeness cost penalizes the AV for getting close to pedestrians. In these equations, $Q_a$, $Q_p$, and $Q_{ED}$ are the weight coefficients.

\begin{equation}\label{eq:J_u}
    J_k^a(a_k) = a_k^\top {Q_a} a_k
\end{equation}

\begin{equation}\label{eq:J_p}
    J_k^p(p_k^{AV}) = Q_p  \left(\frac{\lVert p_k^{AV} - p_{goal} \rVert}{\lVert p_0^{AV} - p_{goal}\rVert} \right)^2
\end{equation}

The pedestrian closeness cost is defined as the sum of the inverse of the Euclidean distances between the AV and all pedestrians within the AV's field of view.

\begin{equation}\label{eq:peds}
    J_k^{ED}(p_k^{AV},s_k^{1:n}) = Q_{ED} \sum_{i=1}^{n} \frac{1}{d_{i,k}^{ED}(p_k^{AV}, s_k^{i})^2}
\end{equation}

Finally, within the terminal cost, the AV is penalized if it fails to reach the goal position by the end of the prediction horizon. The prediction horizon for the MPC optimization problem is determined by the prediction length of the predictor, which is currently set to 6 steps.

\begin{equation}\label{eq:terminal}
    J_K(p_K^{AV}) = J_k^p(p_K^{AV}) |_{k=K}
\end{equation}

This paper addresses the optimization problem outlined in Eq. (\ref{eq:MPC}) by integrating two distinct constraints for collision avoidance adapted from \cite{akhtyamov2023social}: the uncertainty-unaware and uncertainty-aware constraints. Under the uncertainty-unaware constraint, Eq. (\ref{eq:const1}) establishes a minimum threshold for the Euclidean distance between the AV and each pedestrian, utilizing the mean of the predicted distribution of pedestrian positions within the prediction horizon. Conversely, in the uncertainty-aware version, a minimum threshold $\delta$ is set for the collision probability, considering both the covariance of predicted future pedestrian positions and their means. This is translated into a minimum Mahalanobis distance (MD), as stated in Eq. (\ref{eq:const2}) \cite{akhtyamov2023social}.

\begin{equation}\label{eq:const1}
    d_{k,i}^{ED}(p_k^{AV},\hat{s}_k^{i}) \geq (r^{ped} + r^{AV} + PS)
\end{equation}

\begin{equation}\label{eq:const2}
    d_{k,i}^{MD}(p_k^{AV},\hat{s}_k^{i}, \hat{\Sigma}_{k}^{i})^2 \geq -2 \ln(\sqrt{\det(\pi\hat{\Sigma}_k^{i})} \frac{\delta}{V_s})
\end{equation}

\section{Experiments}
\subsection{Simulation Environment}

The majority of crowd navigation algorithms undergo training in a simulation environment where humans are controlled by the Optimal Reciprocal Collision Avoidance (ORCA) model \cite{chen2019crowd, chen2020relational, liu2023intention, chen2017decentralized}. While this simulation environment can be effective for applications involving small mobile robots in pedestrian crowds, our focus in this study is on the navigation of a low-speed autonomous vehicle (AV) in shared spaces with pedestrians. Modelling the heterogeneous interactions in such environments, particularly distinguishing between pedestrian-vehicle and pedestrian-pedestrian interactions, poses a challenge within the ORCA simulation framework. Moreover, the assumption that pedestrians consistently adhere to the ORCA model and actively participate in collision avoidance, as implied by taking half the responsibility, is not always valid for instance in cases involving distracted and non-cooperative pedestrians.

Hence, we employ our custom-built simulation environment, developed based on real-world HBS dataset that has captured pedestrians' and vehicles' trajectories in a shared space \cite{pascucci2017discrete,pascucci2021dataset}. This approach allows us to train our AV in an environment that simulates real pedestrian behaviour. The same dataset is utilized to train the data-driven pedestrian trajectory prediction module.

From this dataset, we extracted 310 scenarios corresponding to the 331 vehicles initially present. Some vehicles that appeared toward the end of the dataset were excluded due to their short and insufficiently captured trajectories. Each extracted scenarios is dedicated to one of the existing vehicles, tracing its trajectory from the initial frame upon entering the scene until it exits. The vehicle's positions at the beginning and end of each scenario are designated as the starting and goal positions within our simulation environment. The 310 scenarios are split into training, testing, and validation sets, with proportions of 64\%, 20\%, and 16\%, respectively.

The behaviours of pedestrians in each scenario are simulated based on real-world data, while the actions of the AV are governed by the DRL policy network. The extracted scenarios involve highly interactive situations, requiring the AV to decelerate to yield to pedestrians or accelerate to pass a pedestrian for avoiding collisions. In the most crowded scenarios, the overall scene can include up to 60 pedestrians over an approximately 1200 $m^2$ area (density of $\approx 0.05\text{ pedestrian}/m^2$). The simulation environment, integrated into a gym environment, has been made publicly available as a benchmark for autonomous vehicle crowd navigation training\footnote{Our source code, the gym environment, and the trained model are available at: \href{https://github.com/Golchoubian/UncertaintyAware_DRL_CrowdNav}{https://github.com/Golchoubian/UncertaintyAware\_DRL\_CrowdNav}}.

\subsection{Experiment setup}

\subsubsection{Baselines and ablation models}

Regarding the data-driven pedestrian trajectory predictor, to demonstrate the advantages of incorporating an uncertainty-aware loss function for more precise predictions of the distribution of pedestrians' future positions, we contrast the original Polar Collision Grid (PCG) model, trained exclusively with the negative log likelihood loss \cite{golchoubian2023polar}, with the same model trained using the updated loss function detailed in Eq. \ref{eq:loss}. We refer to our modified predictor model as Uncertainty-Aware PCG (UAW-PCG).

\textbf{Ablation models for DRL navigation method:} For the entire integrated DRL navigation algorithm, we compare five models. Unless otherwise specified, all models are trained with the speed-independent penalty for danger cases. The first model (DRL - No pred) does not utilize the prediction output in DRL training. The second model (DRL - No pred, speed depend. danger penalty) also omits prediction but is trained with a speed-dependent danger penalty to better capture pedestrians' discomfort at higher vehicle speeds. The third model (DRL - PCG pred) employs the prediction output of the Polar Collision Grid predictor, trained exclusively with a Negative Log Likelihood (NLL) loss function. The fourth model (DRL - UAW-PCG pred) utilizes the prediction output of the uncertainty-aware Polar Collision Grid. Finally, the last model (GT pred) represents a hypothetical model with access to the ground truth future prediction, serving as an upper bound for the performance of models using deterministic prediction outputs. Additionally, we have included the output of all our measures on the trajectory of the actual human-driver for the ego vehicle in the scenario, as per the real dataset (Human-driver).

\textbf{MPC baseline method for social navigation:} We employ three versions of the MPC-based social navigation model proposed in \cite{akhtyamov2023social} and detailed below.

MPC - min dist. hard const.: This MPC model, known as ED-MPC-EDC in \cite{akhtyamov2023social}, incorporates a hard constraint on the minimum Euclidean distance between the AV and each pedestrian. This constraint ensures that the distance exceeds the sum of their two radii and the pedestrian's personal space. 

MPC - min dist. soft const.: This second MPC model, referred to as ED-MPC-AEDC in \cite{akhtyamov2023social}, represents an adaptive version of the first model. In this case, the constraint for avoiding intrusion into the PS of pedestrians is made soft by introducing a slack variable as part of the control inputs to be minimized. Simultaneously, we ensured the model upholds a hard constraint for physical collision avoidance by incorporating an upper bound on the slack variable.

MPC - min collision porb. hard const.: This third MPC model, denoted as ED-MPC-MDC in \cite{akhtyamov2023social}, is an uncertainty-aware MPC. It imposes a hard constraint on keeping the collision probability with the current and predicted position distribution of pedestrians below a threshold $\delta$ (set at 10\% here) as per Eq. (\ref{eq:const2}). 

MPC - min collision prob. hard const. with PCG: This model, is similar to the previous one (MPC - min collision porb. hard const.) but employs the original PCG predictor, instead of the uncertainty-aware predictor. This model aims to showcase the effectiveness of the UAW-PCG prediction model over the original PCG, not only within the context of the DRL model but also for an MPC motion planner that takes prediction uncertainties into account.

\subsubsection{Evaluation metrics}

Given our pedestrian trajectory predictor's emphasis on generating less overconfident predictions by improving the estimation of the covariance matrix for future positions, we assess our predictor module using the measures recommended in \cite{ivanovic2022propagating} as follows:

\begin{itemize}
    \item Average Displacement Error (ADE): Average Euclidean distance between the predicted and the ground truth position over the prediction horizon.
    \item Final Displacement Error (FDE): Average Euclidean distance between the predicted and the ground truth position at the final step of the prediction horizon.
    \item Negative Log Likelihood (NLL): Mean NLL of the ground truth future positions under the predicted distribution outputted by the predictor model.
    \item Delta Empirical Sigma Values ($\Delta ESV_i$): The difference between the empirical proportion of the ground truth positions residing in the i-th standard deviation (e.g., 1$\sigma$, 2$\sigma$, 3$\sigma$) of the anticipated distribution and the same proportion derived from an ideal Gaussian ($\Delta ESV_i: \sigma_{pred,i}-\sigma_{ideal,i}$, where $\sigma_{ideal,1} \approx 0.39$, $\sigma_{ideal,2} \approx 0.86$, $\sigma_{ideal,3} \approx 0.99$). Negative and positive values of $\Delta ESV_i$ indicate overconfident and under-confidence in the predicted distribution accordingly\cite{ivanovic2022propagating}.  
\end{itemize}

\textbf{Evaluation metrics for DRL navigation algorithm: } To assess the trained DRL navigation policy, we incorporate measures for both navigation efficiency and social aspects related to pedestrian comfort. Concerning navigation efficiency metrics, we provide information on the proportion of test set scenarios that successfully reach the goal (Success), collide with any pedestrian (Collision), or experience a timeout (Timeout).

Regarding the social aspects, we employ the measures of intrusion ratio, minimum intrusion distance, and intrusion speed. Intrusion ratio is defined as the percentage of time throughout the entire navigation duration in which the AV intrudes into the personal space of any pedestrian. Furthermore, we define minimum intrusion distance as the smallest distance recorded during these intrusion instances, and intrusion speed as the AV's speed at that distance. We report the mean and standard deviation of these measure across the scenarios in the test set.

\subsection{Implementation Details}\label{imple}

We maintained the network architecture of the DRL model consistent with \cite{liu2023intention}. Pedestrians are represented by circles with a radius of 0.3 meters and a personal space of 1 meter beyond their physical dimension. The AV is modelled with a circle of a 1-meter radius. The maximum speed of the AV is set at 15 km/h, slightly below the allowable speed limit of 20 km/h in the shared space area where the HBS dataset was collected \cite{pascucci2017discrete}, from which we extracted our scenarios. Additionally, we limit the heading change of the AV output from the policy to 0.2 rad/sec. The simulation environment operates with a time step of 0.5 seconds, equivalent to the HBS dataset's frame rate of 2 fps. Each scenario in the simulation environment is configured with a maximum allowable time of 15 seconds more than the total time for the driver to reach the goal point in the real dataset. This designated time serves as the timeout for the DRL training.

The reward for reaching the goal and the penalty for collisions, denoted as $r_g$ and $r_c$ in Eq. \ref{eq:R}, are set to +10 and -20, respectively. The AV is assumed to have a range of view extending up to 15 meters. Lastly, we have chosen the minimum collision probability threshold, $\delta$, to be 0.1. All DRL models undergo training for $4 \times 10^7$ time steps. The initial learning rate is set to $4 \times 10^{-5}$ and undergoes linear decay after $5000$ time steps.

\section{Results and Discussion}

\begin{table} 
\centering
\Huge
\caption{Pedestrian trajectory prediction results. Bolded is the best} \label{tab:Pred}
\resizebox{\columnwidth}{!}{\begin{tabular}{|l|c|c|c|c|c|c|}
\hline
\textbf{Model} & \textbf{ADE (m)} & \textbf{FDE (m)} & \textbf{NLL} & $\bm{\Delta ESV_1}$ & $\bm{\Delta ESV_2}$ & $\bm{\Delta ESV_3}$ \\
\hline
PCG & 0.343\LARGE{$\pm$0.15} & 0.753\LARGE{$\pm$0.33} & 6.323\LARGE{$\pm$14.88} & \textbf{-0.231} & -0.464 & -0.415\\
\hline
UAW-PCG & \textbf{0.333}\LARGE{$\pm$0.14} & \textbf{0.732}\LARGE{$\pm$0.32} & \textbf{0.537}\LARGE{$\pm$3.06} & 0.440 & 0\textbf{.089} & \textbf{-0.012}\\
\hline
\end{tabular}}
\end{table}

\begin{figure}[t!]
\centering
\subfloat[PCG]{\includegraphics[width=3.2in]{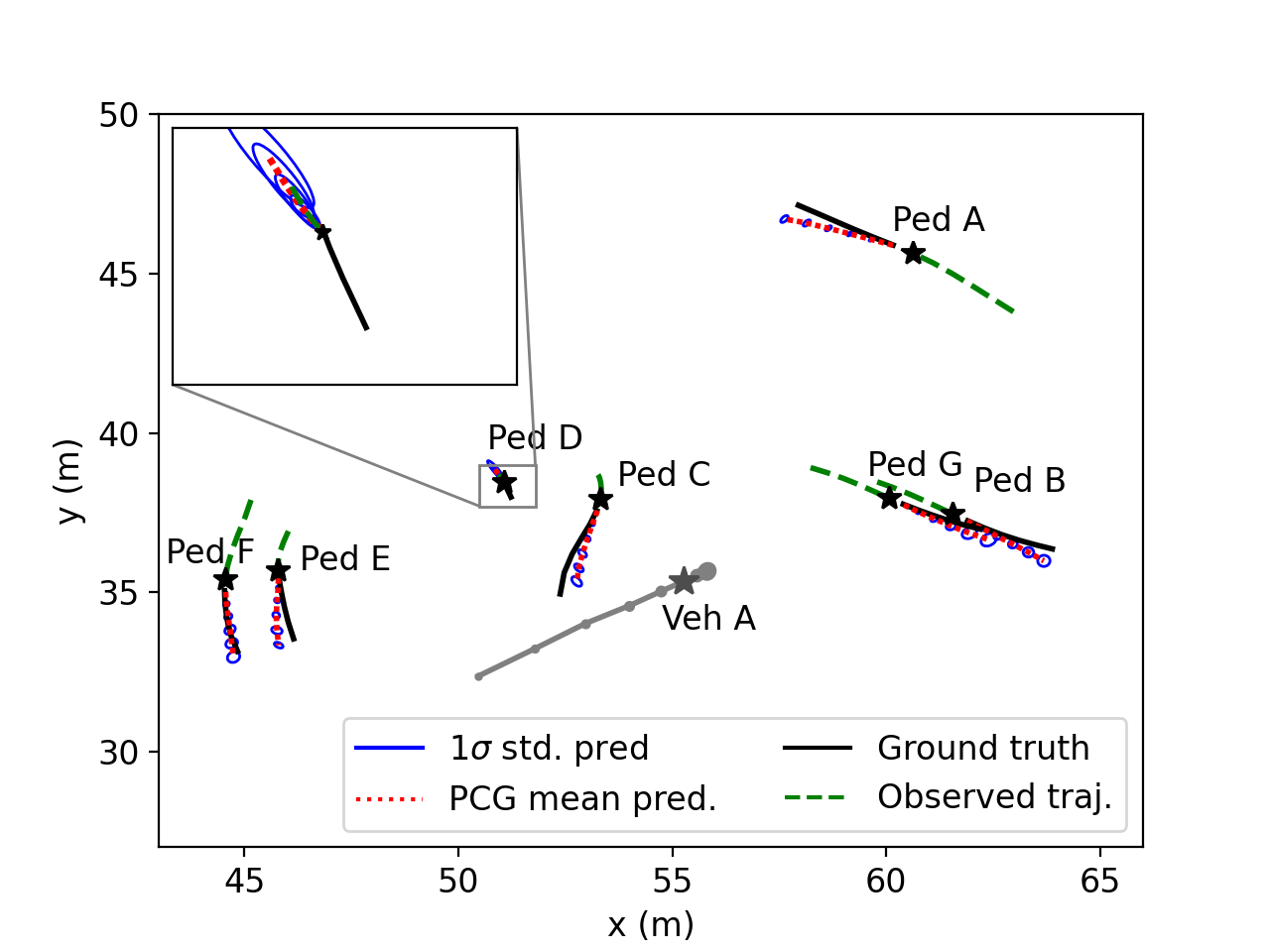}%
\label{fig:874PCG}}
\vspace{-1.3em} 
\hfil
\subfloat[UAW-PCG]{\includegraphics[width=3.2in]{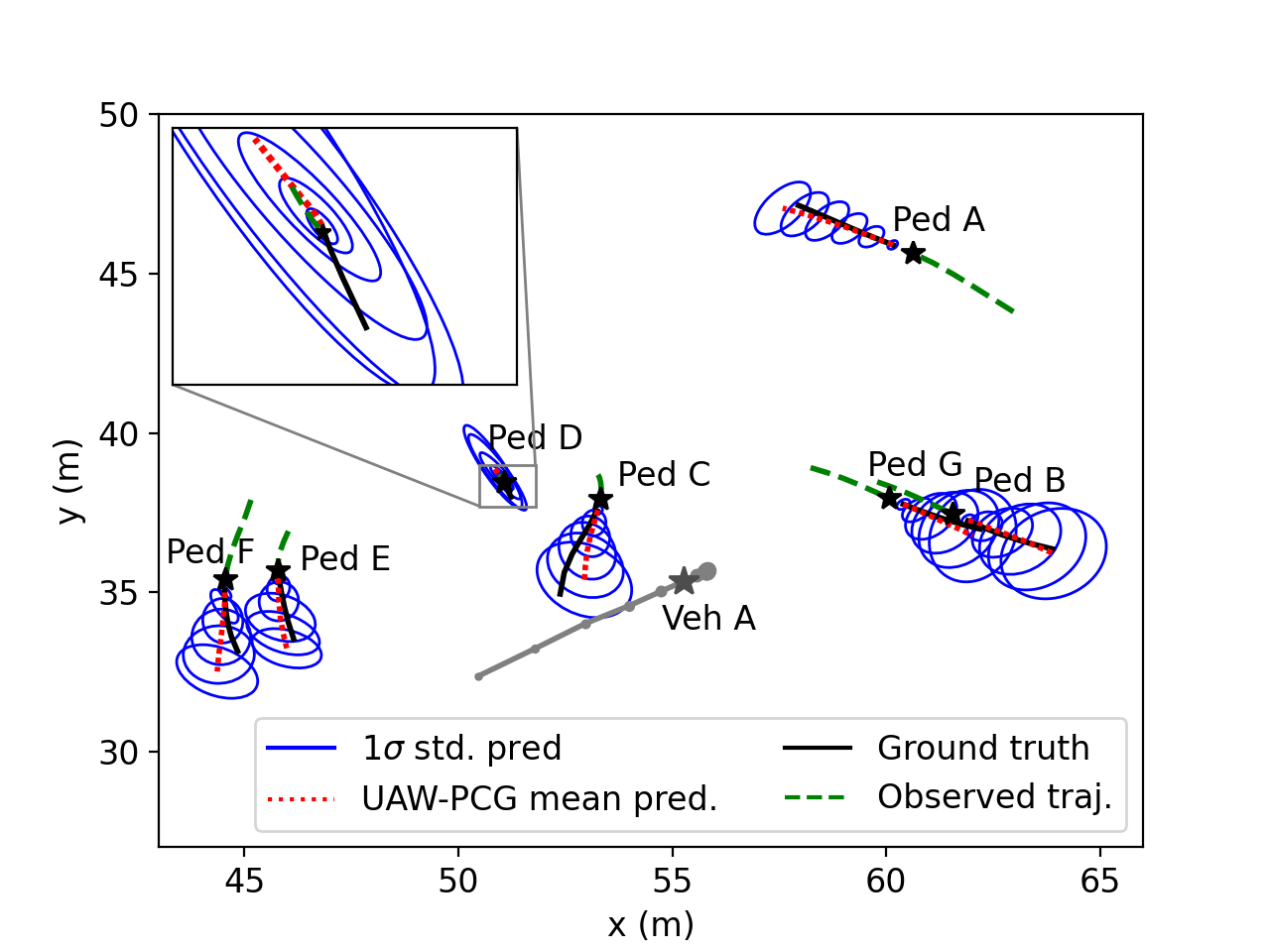}%
\label{fig:874UAWPCG}}
\caption{Predicted trajectories of (a) the polar collision grid (PCG) model trained with only the negative log likelihood loss, compared to (b) the same model trained with an additional uncertainty loss, referred to as the Uncertainty-Aware PCG (UAW-PCG). The blue circles represent the $1\sigma$ standard deviation of the predicted distribution for the future position around the estimated mean (depicted in red). Agents positions at the current time steps are marked with a star, and larger markers along the vehicle's trajectory highlight positions at later times. Here ``Ped" and ``Veh" stand for pedestrian and vehicle respectively.}
\label{fig:874}
\end{figure}

\begin{table*} 
\centering
\Huge
\caption{DRL navigation and baseline results. Bold is the best. Underline is the second best} \label{tab:DRL}
\resizebox{\textwidth}{!}{\begin{tabular}{|l|c|c|c|c|c|c|c|c|l|}
\hline
\textbf{Model} & \textbf{Success} & \textbf{Collision} & \textbf{Timeout} & \parbox{3cm}{\centering \textbf{Nav.} \\ \textbf{ time($s$)}} & \parbox{4cm}{\centering \textbf{Path} \\ \textbf{length ($m$)}} & \parbox{4cm}{\centering \textbf{Intrusion ratio (\%)}} & \parbox{4cm}{\centering \textbf{Intrusion dist. ($m$)}} & \parbox{5cm}{\centering \textbf{Intrusion speed ($m/s^2$)}} \\
\hline
MPC - min. dist. hard const. & 0.84 & 0.07 & 0.09 & 15.29\LARGE{$\pm$5.68} & 47.94\LARGE{$\pm$6.78} & 1.55\LARGE{$\pm$4.77} & 0.72\LARGE{$\pm$0.27} & 2.47\LARGE{$\pm$1.88} \\
\hline
MPC - min dist soft const. & 0.84 & 0.07 & 0.09 & 15.24\LARGE{$\pm$6.02} & 47.81\LARGE{$\pm$6.71} & 2.02\LARGE{$\pm$5.63} & \textbf{0.79}\LARGE{$\pm$0.25} & 2.79\LARGE{$\pm$1.76}\\
\hline
MPC - min collision prob. hard const. & 0.88 & 0.05 & 0.07 & 15.49\LARGE{$\pm$5.82} & 48.73\LARGE{$\pm$8.16} & \underline{0.07}\LARGE{$\pm$0.50} & 0.02\LARGE{$\pm$0.00} & 4.16\LARGE{$\pm$0.00}\\
\hline
MPC - min collision prob. hard const. with PCG & 0.84 & 0.09 & 0.07 & 14.88\LARGE{$\pm$5.13} & 47.67\LARGE{$\pm$6.65} & \textbf{0.00}\LARGE{$\pm$0.00} & - & - \\
\hline
DRL - No pred & 0.84 & 0.07 & 0.09 & \textbf{13.41}\LARGE{$\pm$3.49} & 46.25\LARGE{$\pm$6.07} & 3.79\LARGE{$\pm$5.75} & 0.68\LARGE{$\pm$0.27} & 2.12\LARGE{$\pm$1.49}\\
\hline
DRL - No pred, speed depend. danger penalty & \underline{0.90} & \textbf{0.02} & 0.09 & 15.24\LARGE{$\pm$5.15} & 46.65\LARGE{$\pm$6.52} & 2.35\LARGE{$\pm$4.04} & 0.68\LARGE{$\pm$0.26} & \textbf{1.84}\LARGE{$\pm$1.44}\\
\hline
DRL - PCG pred & \underline{0.90} & 0.05 & \underline{0.05} & \underline{14.50}\LARGE{$\pm$4.96} & \textbf{45.41}\LARGE{$\pm$5.69} & 1.80\LARGE{$\pm$3.67} & 0.64\LARGE{$\pm$0.24} & \underline{1.99}\LARGE{$\pm$1.74}\\
\hline
DRL - UAW-PCG pred (Ours) & \textbf{0.95} & \underline{0.03} & \textbf{0.02} & 14.62\LARGE{$\pm$4.40} & \underline{46.06}\LARGE{$\pm$6.00} & 3.11\LARGE{$\pm$5.38} & \underline{0.74}\LARGE{$\pm$0.21} & 2.00\LARGE{$\pm$1.28}\\
\hline
\hline
DRL - GT pred & 0.95 & 0.02 & 0.03 & 14.47\LARGE{$\pm$4.16} & 45.60\LARGE{$\pm$5.74} & 2.24\LARGE{$\pm$3.56} & 0.68\LARGE{$\pm$0.25} & 2.62\LARGE{$\pm$1.39}\\
\hline 
Human-driver & 1.00 & 0.00 & 0.00 & 16.10\LARGE{$\pm$5.57} & 45.83\LARGE{$\pm$6.59} & 2.54\LARGE{$\pm$3.93} & 0.62\LARGE{$\pm$0.29} & 2.07\LARGE{$\pm$1.66} \\ 
\hline
\end{tabular}}
\end{table*}

\subsection{Uncertainty-aware Trajectory Predictor} \label{res-pred}

\subsubsection{Quantitative results}

Table \ref{tab:Pred} presents the results of the Polar Collision Grid model, trained with our proposed combination loss, in comparison to the same model trained solely with the NLL loss, as done in \cite{golchoubian2023polar}.

Upon comparing the two trained models, it is evident that the model trained exclusively with NLL exhibits overconfident predictions, as indicated by its negative $\Delta ESV_i$ values (e.g., $\Delta ESV_3$ = -0.415). In contrast, the model trained with the combination loss yields a $\Delta ESV_3$ value closer to zero (-0.012), which is considered ideal. Additionally, larger positive values are observed for $\Delta ESV_1$ indicating underconfidence for the uncertainty-aware predictor. However, underconfidence in predictions is preferable compared to overconfident predictions when using the predictor results for planning. This approach provides a wider safety margin and promotes a more conservative behaviour.

Notably, the model exclusively trained with NLL loss demonstrates a higher standard deviation in the NLL loss measure on the test set when compared to the model trained with the combination loss. This difference also highlights the overconfident predictions of the original model. Despite occasionally achieving very low NLL values towards the lower end of the NLL distribution over the test cases (6.323 $\pm$ 14.88) the model struggles to generalize effectively to unseen data in the test set.

Therefore, the incorporation of the uncertainty loss has proven beneficial, enabling the model to better estimate the uncertainty (covariance matrix) of its predicted distribution for the future positions of pedestrians. This enhancement is accompanied by a slight improvement in the Average Displacement Error (ADE) and Final Displacement Error (FDE) measures, with ADE decreasing by 1 cm and FDE decreasing by 2.1 cm. In Table \ref{tab:Pred}, ADE and FDE values are reported for the predicted mean to facilitate a fair comparison. Therefore, these values may differ slightly from the original paper, which reported metrics on the best 20 samples from the predicted distribution.

\subsubsection{Qualitative results}

To conduct a qualitative analysis of the predicted trajectories and visually assess the overconfidence prediction of the original model, we have generated visualizations of the predictions for a sample case from the test set, as depicted in Fig. \ref{fig:874}. This figure illustrates that, despite potential deviations between the predicted mean and the ground truth in certain cases, the Uncertainty-Aware PCG (UAW-PCG) assigns a probability to the occurrence of the ground truth. In contrast, the original model tends to overlook this probability, especially in scenarios involving stationary pedestrians who may unexpectedly initiate movement in subsequent time steps, like the case of Ped D in Fig. \ref{fig:874}. For such cases the UAW-PCG excels in anticipating higher uncertainties in the next time step, capturing the likelihood of sudden movements. Consequently, for enhancing pedestrian safety in Deep Reinforcement Learning (DRL) training for collision avoidance, we rely on predictions from the UAW-PCG.

Another notable observation is that the Uncertainty-Aware PCG (UAW-PCG) model grasps the pattern of uncertainty propagation across the prediction horizon. It consistently predicts distributions with higher standard deviations for time steps farther into the future. This aligns with the correct understanding that prediction uncertainty naturally increases for longer horizons. This crucial pattern is not distinctly evident in the overconfident predictions of the model trained solely with Negative Log Likelihood (NLL).

\begin{figure*}[!t]
\centering

\subfloat[DRL - UAW-PCG pred model]{\includegraphics[width=7.0in]{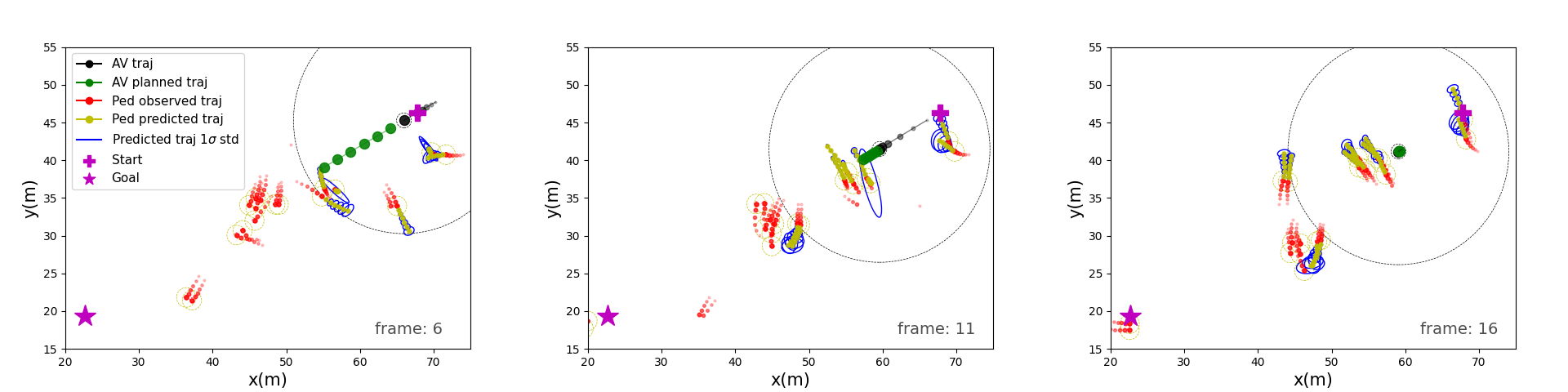}} \vspace{-1.0em}

\subfloat[DRL - No pred, speed depend. danger penalty model]{\includegraphics[width=7.0in]{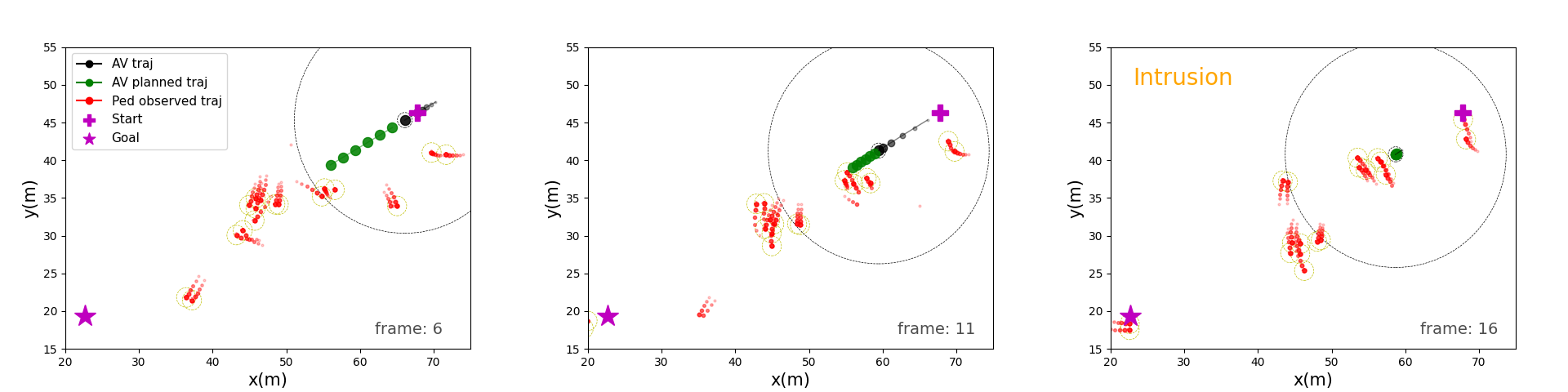}} \vspace{-1.0em}

\subfloat[DRL - No pred model]{\includegraphics[width=7.0in]{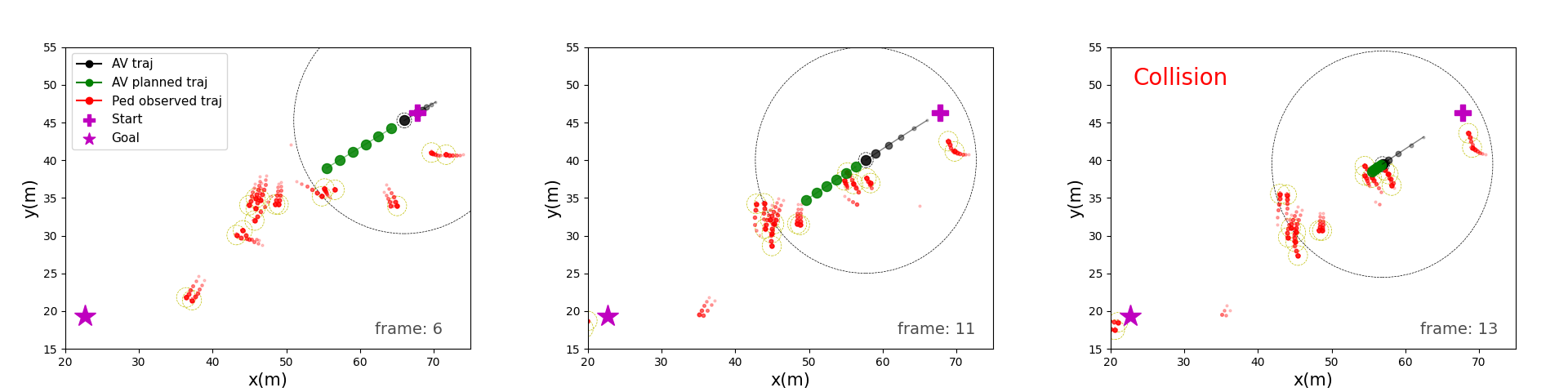}} \vspace{-1.0em}

\subfloat[MPC - min collision prob. hard const. model]{\includegraphics[width=7.0in]{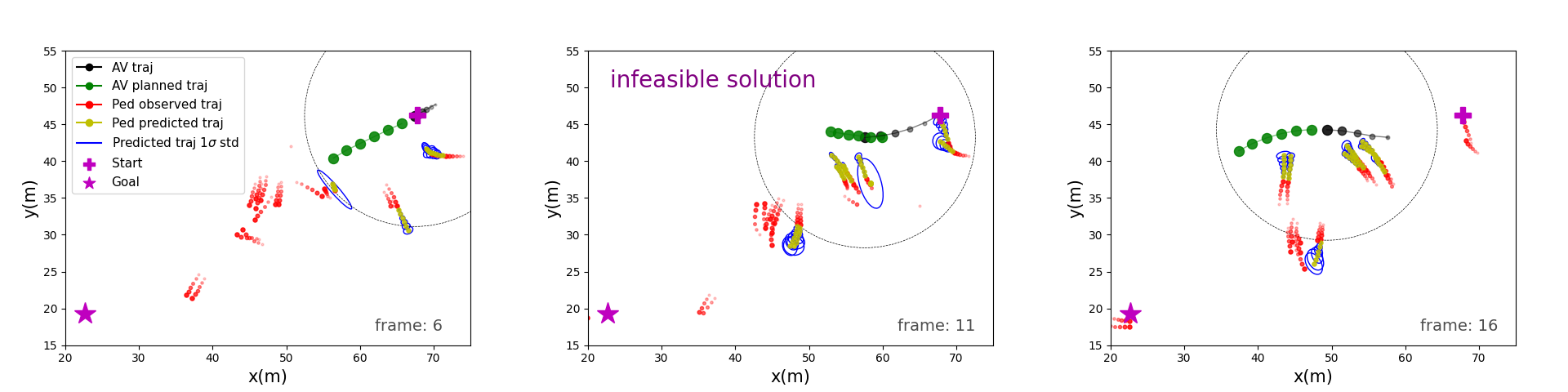}}

\caption{Comparison of various model trajectories in identical test scenarios. The figure illustrates an autonomous vehicle (in black) navigating through a crowd of pedestrians (in red) towards a goal indicated by a star. It depicts the trajectories of both the AV and the pedestrians for the past 6 frames. A dashed yellow line encircles the current position of each pedestrian, representing their personal space. For models incorporating predictions and uncertainty, the mean trajectory and 1$\sigma$ standard deviation of the pedestrians' future paths over the next 6 frames are depicted by yellow and blue ellipses, respectively. The AV's maximum sensor detection range is shown as a dashed black circle around its current position, with predictions being made only for pedestrians within this range. The detailed progress of each model in this scenario can be found in the attached video.}
\label{fig:295}
\end{figure*}

\begin{figure*}[!t]
\centering

\subfloat[DRL - UAW-PCG pred model]{\includegraphics[width=7.3in]{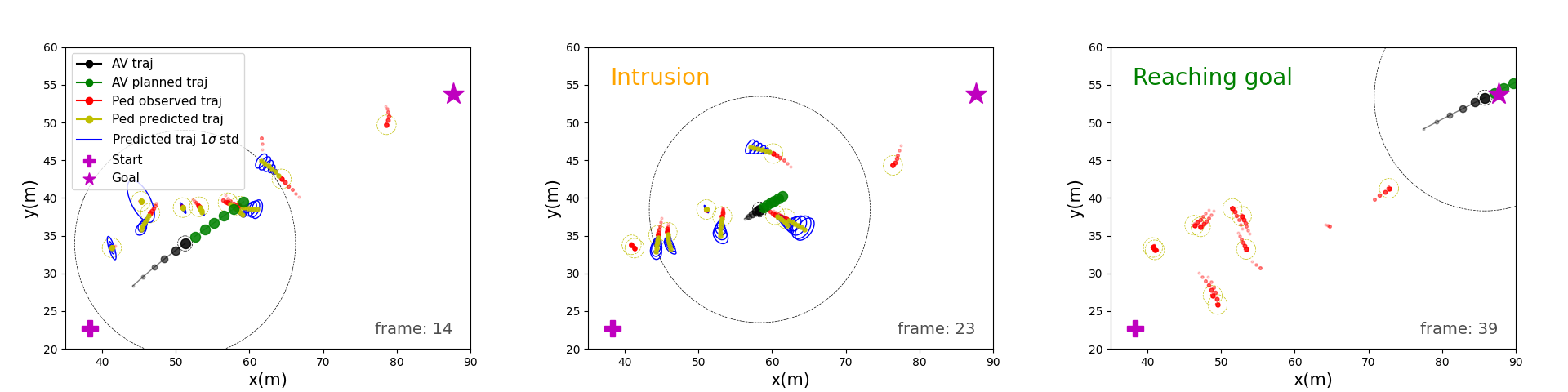}} \vspace{-1.0em}

\subfloat[DRL - PCG pred model]{\includegraphics[width=7.3in]{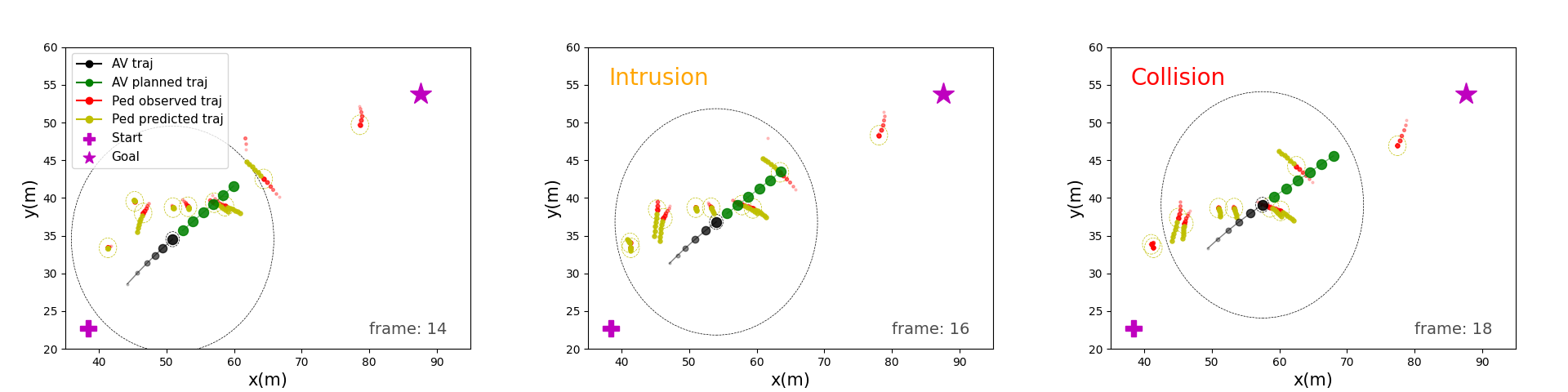}}

\caption{Comparison of the Autonomous Vehicle's (AV) trajectory behaviour in an identical test scenario. This figure presents the trajectory of the AV using our proposed Deep Reinforcement Learning (DRL) model, which accounts for prediction uncertainties (UAW-PCG pred), against the baseline model that does not consider such uncertainties (PCG pred). The trajectories are illustrated over three consecutive frames.}
\label{fig:300}
\end{figure*}

\subsection{DRL motion planner}
\subsubsection{Quantitative results}
The results of the trained DRL motion planner and its ablations as well as the MPC baseline model are presented in Table \ref{tab:DRL}.

\textbf{Effectiveness of deep reinforcement learning algorithm:} The DRL models demonstrate superior performance compared to MPC models, exhibiting an overall higher success rate and a lower collision rate. Additionally, the DRL model tends to follow more direct paths to the goal, resulting in a reduced path length and lower navigation times in comparison to MPC models (refer to Table \ref{tab:DRL}). The average computational time required for each model to determine an optimal action at every time step is detailed in Table \ref{tab:Compute_time}. The findings reveal that the DRL model operates more than twice as fast as the MPC model. This is particularly crucial for real-time implementation in crowded spaces, where satisfying a multitude of constraints (one for each visible pedestrian) can lead to extended convergence times for MPC, or even infeasibility in highly dense spaces. Consequently, the DRL model, which offloads computation time to the training phase and exhibits faster performance during inference, holds a distinct advantage in terms of computational efficiency.

Collisions in the MPC model arise either when it encounters infeasible solutions in crowded scenarios or when the predicted collision probability falls below the set threshold in the constraint for the MPC model with a minimum collision probability constraint (MPC - min col. prob. hard const.). The latter cause of collisions can potentially be linked to the fact that Eq. (\ref{eq:P_Col}) is an approximation and is subject to some error. Introducing a soft constraint to relax the hard constraint associated with avoiding pedestrians' personal space in the (MPC, min dist. soft. const.) model can mitigate the occurrence of infeasible solutions. However, this comes at the cost of increased intrusion into pedestrians' personal space, as reported in Table. \ref{tab:DRL}.

The MPC model with a hard constraint defined on the minimum collision probability exhibits the lowest collision rate of 0.05 among all MPC models. This highlights that the MPC model which considers the uncertainty of pedestrians' next states, is better equipped to avoid intrusion into pedestrians' personal space (as indicated by the intrusion ratio in Table. \ref{tab:DRL}), ultimately resulting in a reduced collision rate. However, the success of this MPC model depends on accurate uncertainty predictions. This is evident in the higher collision rate observed in a similar MPC model that, instead of utilizing the predictions of the UAW-PCG, relies on the overconfident predictions of the original PCG (MPC - min collision prob. hard const. with PCG).

\textbf{Effectiveness of speed dependent danger penalty:} In the two cases of the DRL model where no prediction is used, the model trained with a speed-dependent danger penalty exhibits a significantly reduced collision rate and intrusion ratio compared to the model trained with a speed-independent danger penalty in the reward function (Table. \ref{tab:DRL}). Additionally, the speed of the AV during the intrusion of pedestrians' personal space for the DRL model trained with a speed-dependent penalty is lower than the speed of the AV in intrusion cases in the baseline model (DRL - No pred). Given that the only difference between these two models lies in the danger penalty, the results suggest that incorporating the impact of the AV's speed during intrusion on people's discomfort into the reward function contributes to learning a policy that better respects people's personal space and reduces collision occurrences in the simulation.

\textbf{Effectiveness of prediction:} Consistent with the findings in \cite{liu2023intention}, incorporating pedestrians' predicted trajectories during the training of the DRL model has a notable impact. This is evident in the increased success rate, reduced collision rate and timeout rate, as well as a lower intrusion ratio, as detailed in Table. \ref{tab:DRL}. The comparison between the DRL model that does not consider prediction (DRL - No pred) and the one with prediction (DRL, PCG pred) and (DRL, UAW-PCG pred) highlights these improvements.

However, this enhancement comes at the expense of increased computational time for generating predictions, as shown in Table \ref{tab:Compute_time}. Nonetheless, the computational time remains significantly below the 0.5-second time step at which the simulation operates.

\begin{figure}[b!]
  \centering
  \includegraphics[width=0.8\columnwidth]{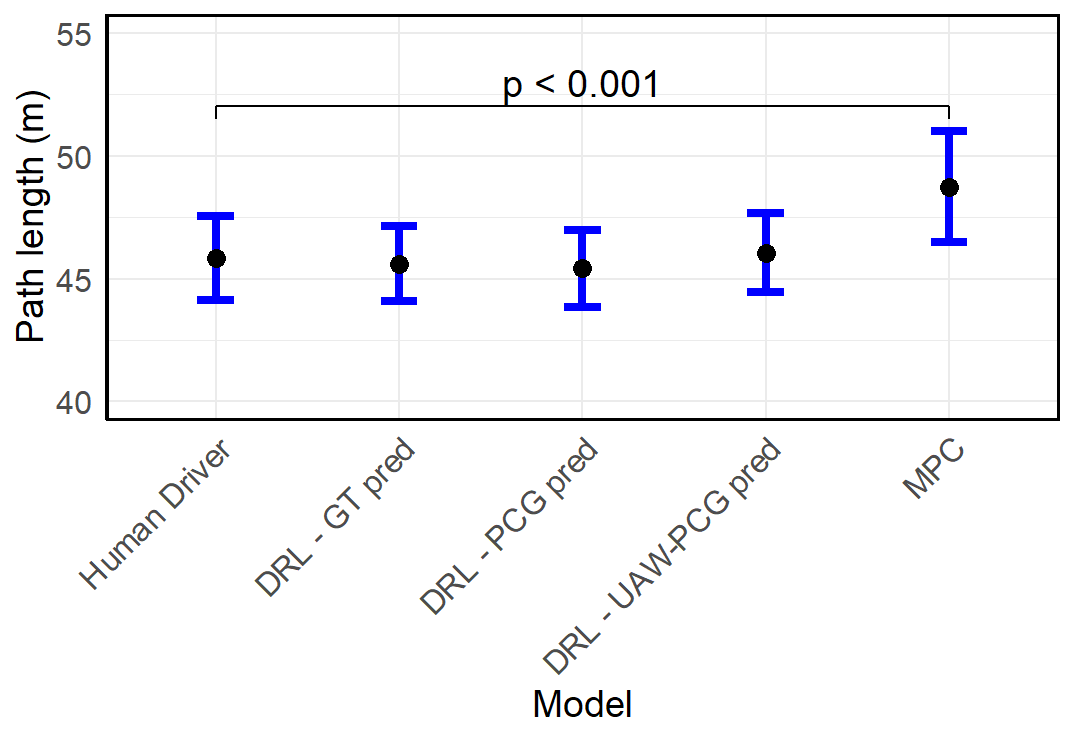}
  \caption{Mean and 95\% confidence interval of difference model's path length over the test set scenarios}
  \label{fig:path_len}
\end{figure}

\textbf{Effectiveness of the uncertainty-aware DRL model: } Our proposed model, which considers uncertainties in pedestrians' predicted trajectories (DRL - UAW-PCG), exhibits a 40\% decrease in collision rate and 60\% decrease in timeout rate, along with an overall 5.5\% increase in the success rate, compared to the baseline model that neglects prediction uncertainties (DRL - PCG pred), as detailed in Table. \ref{tab:DRL}. Furthermore, our uncertainty-aware prediction model is observed to maintain a greater distance from pedestrians when intruding into their personal space. The hypothetical DRL model assuming access to ground truth predictions without considering prediction uncertainty outperforms the same model relying on the predicted mean of the PCG module. This observation emphasizes that the performance degradation in the uncertainty-unaware model (DRL - PCG pred) is attributed to predictions deviating from ground truth trajectories, emphasizing the importance of considering uncertainties. Consequently, our model, which accounts for these uncertainties, demonstrates performance closer to the hypothetical model with access to ground truth (GT pred), and even surpasses it on certain safety metrics such as intrusion distance and intrusion speed. This superiority is derived from its more cautious behaviour, stemming from reliance on the predicted distribution of pedestrian future positions rather than single future position points.

\textbf{Comparison to human driver trajectory:} The average path length of our proposed model is slightly higher than that of the human driver in the dataset, while achieving a lower navigation time (Table. \ref{tab:DRL}). This indicates that the average velocity of the AV in our model was higher, while deviating slightly more from the straight path to the goal compared to the human driver in the same scenario. 
It is important to note that these differences in path length of our DRL model compared to the human-driven trajectory, is not significant when contrasted with the path length of the MPC model. The MPC model's path length is statistically longer than the overall path length of the human driver ($se = 0.548, t = -4.121, p < 0.001 $), as illustrated in Figure \ref{fig:path_len}. This differences is attributed to the MPC model primarily deviating from a straight path for collision avoidance, whereas our model's strategy prioritizes speed adjustment, resulting in a trajectory closer to that of the human driver.

It is important to note that the intrusion ratio in the human driver case is not zero due to the errors introduced by modelling the vehicle with a circle. This results in the simulation environment detecting intrusion from the side when the vehicle passes a pedestrian, despite maintaining an acceptable distance with its rectangular shape in the real world data. The conservative assessment with a circular shape model for the vehicle contributes to this detection of intrusion.

\begin{table}[t]
\centering
\caption{One step average computational time of different models} \label{tab:Compute_time}
\resizebox{\columnwidth}{!}{\begin{tabular}{|l|c|}
\hline
\textbf{Model} & \textbf{Computational time (sec)} \\
\hline
DRL - No pred & 0.01\tiny{$\pm$0.00}\\
\hline
DRL - UAW-PCG pred & 0.04\tiny{$\pm$0.02}\\
\hline
MPC - min collision prob. hard const. & 0.10\tiny{$\pm$0.08}\\
\hline
\end{tabular}}
\end{table}

\begin{table*} 
\centering
\caption{Sensitivity analysis of the reward value for goal achievement ($r_g$). The default value is denoted by (*).
} \label{tab:Sensitivity}
\resizebox{\textwidth}{!}{\begin{tabular}{|l|c|c|c|c|c|c|c|c|l|}
\hline
\textbf{$r_g$} & \textbf{Success} & \textbf{Collision} & \textbf{Timeout} & \textbf{Nav. time($s$)} & \textbf{Path length ($m$)} & \textbf{Intrusion ratio (\%)} & \textbf{Intrusion dist. ($m$)} & \textbf{Intrusion speed ($m/s^2$)} \\
\hline
1 & 0.93 & 0.00 & 0.07 & 15.11\tiny{$\pm$5.02} & 45.66\tiny{$\pm$6.00} & 2.17\tiny{$\pm$4.30} & 0.64\tiny{$\pm$0.22} & 1.75\tiny{$\pm$1.47} \\
\hline
10* & 0.95 & 0.03 & 0.02 & 14.62\tiny{$\pm$4.40} & 46.06\tiny{$\pm$6.00} & 3.11\tiny{$\pm$5.38} & 0.74\tiny{$\pm$0.21} & 2.00\tiny{$\pm$1.28} \\
\hline
100 & 0.90 & 0.07 & 0.03 & 13.28\tiny{$\pm$4.13} & 47.09\tiny{$\pm$6.72} & 4.25\tiny{$\pm$7.17} & 0.62\tiny{$\pm$0.28} & 2.26\tiny{$\pm$1.11} \\
\hline
\end{tabular}}
\end{table*}

\textbf{Reward Sensitivity Analysis}: The values in the reward function were selected based on the literature \cite{liu2023intention,liu2021decentralized} and manually adjusted to enhance convergence and strike a balance between safety and efficiency. While a comprehensive parameter tuning is left for future exploration, we briefly investigate the trained model's sensitivity to the goal achievement reward value ($r_g$), a key parameter in the reward function, while keeping other parameters at their default values (Section \ref{imple}).

With the default value of $r_g$ set to 10, we also trained and evaluated models with 1/10 and 10 times the default value (see Table \ref{tab:Sensitivity}). Reducing $r_g$ to 1 results in zero collision rate at the expense of a slightly reduced success rate due to increased timeouts. This happens because, with a lower $r_g$ compared to the default collision penalty of -20, the model prioritizes collision avoidance over reaching the goal, leading to longer navigation times on average and demonstrating a more conservative behaviour.

Conversely, significantly increasing the reward value for reaching the goal to 100 results in a policy that prioritizes goal achievement at the expense of collision avoidance. In this case, the AV rushes to reach the goal, resulting in shorter navigation times but riskier behaviour, as evidenced by higher intrusion rates, increased speed during intrusions, and shorter intrusion distances in Table \ref{tab:Sensitivity}. This behaviour ultimately leads to higher collision rate and a decrease in the overall success rate. The trends revealed in this sensitivity analysis underscore the critical role of the reward function and how adjusting the relative reward values can assist in achieving an acceptable balance between safety and efficiency.

\subsubsection{Qualitative results}

The behaviours of three DRL models, along with the best MPC model, are compared in the same scenario from the test set in Fig. \ref{fig:295}. The baseline model, which does not use predictions (DRL - No pred), ends up getting too close to a pedestrian and ultimately colliding with it. The model trained with a speed-dependent danger penalty successfully avoids any collision in the same scenario by learning to slow down when approaching pedestrians, yet it still intrudes into a pedestrian's personal space at frame 16. In contrast, our proposed uncertainty-aware DRL model, utilizing future predicted states of pedestrians and the associated uncertainties, successfully reaches the goal in this scenario without intruding into any pedestrian's personal space. This is achieved by predicting high uncertainties for a pedestrian in frame 11, prompting the AV to slow down from that frame.

The comparison between our uncertainty-aware DRL model and the best-performing MPC model in the same scenario is also illustrated in Fig. \ref{fig:295}. Although both models successfully reach the goal, the MPC model encounters infeasibility over a couple of frames including frame 11. Despite not colliding with any pedestrian, this results in a highly unnatural planned path for that time step, posing a potential risk of collision. Another noticeable difference in the output behaviour of these two models is that the MPC model avoids the crowd in front by deviating to one side, while the DRL model achieves collision avoidance through speed adjustment, primarily maintaining a straight line towards the goal. This distinction is supported by the average lower path length of DRL methods compared to the MPC models, as shown in Table \ref{tab:DRL} and Fig \ref{fig:path_len}. This behavioural difference is observed despite both models being constrained to the same upper and lower bounds for speed and heading change, and both models being encouraged to keep the heading change low (heading change control appears in the stage cost of MPC as well as the reward function of the DRL model). Besides, considering the fact that speed adjustment behaviour better aligns with the path taken by the human driver in the same scenario from the real dataset, it is noteworthy that for collision avoidance, speed adjustment has been shown to be perceived safer by people compared to heading change \cite{golchoubian2021social}.

In Fig. \ref{fig:300}, the performance of our uncertainty-aware model (DRL - UAW-PCG pred) is compared with its baseline, the uncertainty-unaware counterpart model (DRL - PCG pred), in another scenario from the test set. By predicting uncertainties associated with pedestrians' future positions, our model maintains a safe distance from pedestrians on its left in this scenario through a gradual deviation to the right. In contrast, the uncertainty-unaware model, relying only on the actual predicted future position, causes the AV to drive too close to the pedestrian, intrude into their personal space and finally colliding with one. In this scenario, our model yields to the pedestrian that the baseline model has collided with. However, due to early acceleration after the stop, the AV in our model slightly intrudes into the pedestrian's personal space from behind after they have passed. Nevertheless, unlike its counterpart, our model manages to reach the goal successfully. This scenario serves as an example of our model's ability to output more cautious behaviour by considering uncertainties associated with pedestrians' future trajectories.

\section{Limitations}

Our proposed method and simulation environment, while effective, have limitations. We utilized a unicycle model for AV representation, suitable for low operational speeds, which is an improvement over other methods using holonomic kinematics \cite{liu2021decentralized, liu2023intention,chen2020relational,chen2019crowd}. However, future assessments could explore more complex kinematics or dynamic models. Additionally, we modelled the AV as a circle in the simulation to facilitate distance calculations, potentially resulting in a larger than necessary distance during side passes due to the vehicle's rectangular shape. This could be addressed in the future by incorporating an elliptical model for the AV's shape.

Our model was exclusively tested on the HBS dataset, primarily due to the limited availability of datasets depicting pedestrian and vehicle trajectories closely interacting within unstructured or shared spaces. Evaluating the generalizability of the learned model can be pursued in future studies as more datasets become accessible.

Lastly, the simulation environment replays realistic human trajectories but lacks responsiveness to various AV behaviours as humans would react in the real world. The trained policy within this environment minimizes reliance on pedestrian cooperation for collision avoidance, reducing the need for pedestrians to adjust paths in real-world implementation. Future work aims to enhance the simulation environment's responsiveness by using a prediction algorithm accounting for the vehicle's interaction effects in a closed-loop form.

\section{Conclusion and future work}

In this work, we proposed integrating prediction uncertainties into policy learning of a model-free deep reinforcement learning algorithm. This algorithm is developed for navigating a low-speed vehicle among pedestrians in an unstructured environment. We employed a data-driven pedestrian trajectory predictor trained with a novel uncertainty-aware loss function, generating uncertainties as a bivariate Gaussian distribution considering pedestrian-AV interaction effects. These predictions, along with their covariance matrix, are inputted into the DRL motion planner, which is trained with a novel reward function. This reward function aims to encourage the vehicle to maintain a safe distance from pedestrians' current positions, minimize collision probability with their predicted future positions, and consider pedestrians' comfort during close approaches. To simulate realistic pedestrian behaviour in presence of vehicles, we trained our model in a simulation environment derived from real-world pedestrian trajectories.

In summary, our contributions include: 1) Integrating prediction uncertainties into a model-free DRL algorithm for AV crowd navigation. 2) Introducing a novel reward function that considers pedestrians' discomfort based on AV distance and speed. 3) Developing a simulation environment for AV crowd navigation using realistic pedestrian behaviours from real-world datasets. 4) Proposing a new loss function for trajectory prediction to mitigate overconfident predictions.

Our findings demonstrate the effectiveness of our uncertainty-aware DRL model in reducing collision and timeout rates while increasing the minimum distance during intrusion. Compared to Model Predictive Control (MPC) models, which serve as an example of optimal control methods for crowd navigation, our model exhibits faster computational times, superior performance, and trajectories closely resembling those of human drivers in similar scenarios. To further enhance our proposed framework, future work could explore incorporating the effect of the approach direction of the AV on pedestrians' comfort, as well as considering passenger comfort constraints. Additionally, modelling the pedestrians' personal space as directional, extending more prominently in their front, could be considered and integrated into the design of the reward function.

\section*{Acknowledgments}
This research was undertaken, in part, thanks to funding from the Canada 150 Research Chairs Program and NSERC.

\bibliographystyle{IEEEtran}
\bibliography{IEEEabrv,mybibfile}

\newpage

\section{Biography Section}

\begin{IEEEbiography}[{\includegraphics[width=1in,height=1.25in,clip,keepaspectratio]{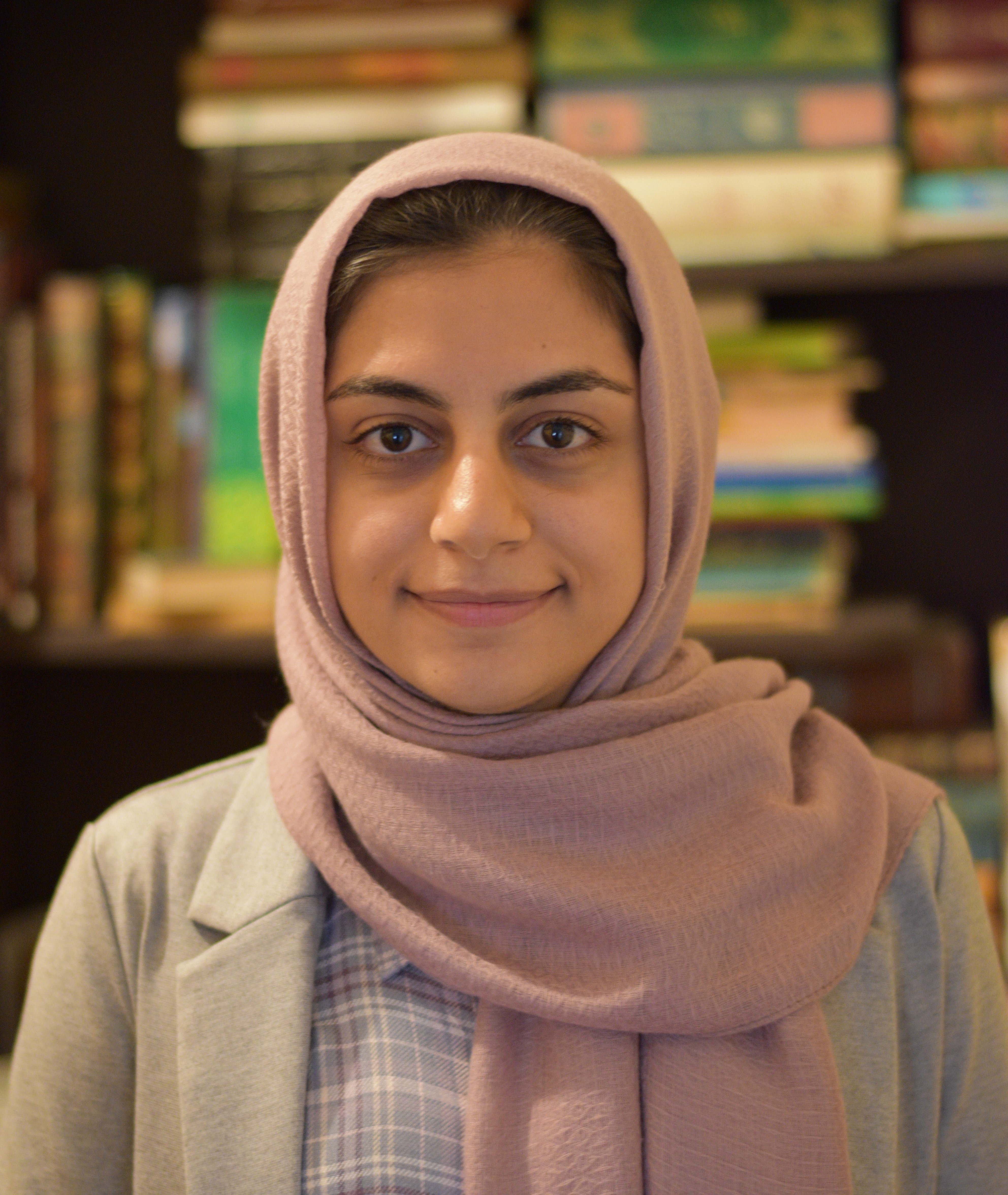}}]{Mahsa Golchoubian} is a Ph.D. candidate at the University of Waterloo’s Systems Design Engineering Department, Canada. She received her B.Sc. and M.Sc. in Aerospace Engineering from Sharif University of Technology, Iran, in 2015 and 2018, respectively. Currently, she is conducting research at the intersection of autonomous navigation, human-robot interaction, and machine learning.
\end{IEEEbiography}

\begin{IEEEbiography}[{\includegraphics[width=1in,height=1.25in,clip,keepaspectratio]{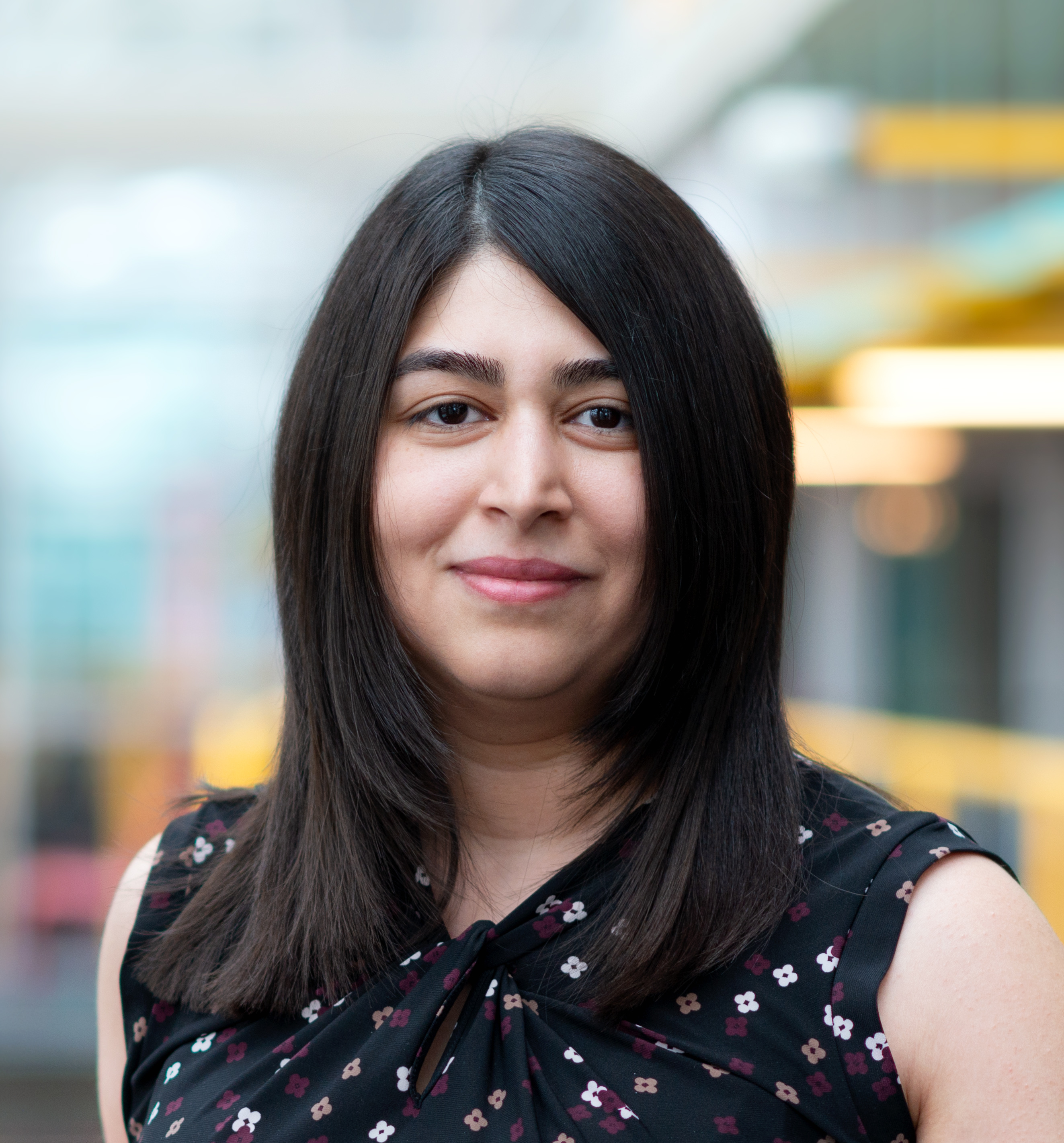}}]{Moojan Ghafurian} is an Assistant Professor in the Department of Systems Design Engineering at the University of Waterloo. She got her PhD from the Pennsylvania State University and was the Inaugural Wes Graham postdoctoral fellow from 2018-2020 at David R. Cheriton School of Computer Science at the University of Waterloo. Her research areas are human-computer/robot interaction, social robotics, affective computing, and cognitive science. Her research explores computational models of how humans interact with systems to inform user-centered design of emotionally and socially intelligent agents in multiple domains.
\end{IEEEbiography}

\begin{IEEEbiography}[{\includegraphics[width=1in,height=1.25in,clip,keepaspectratio]{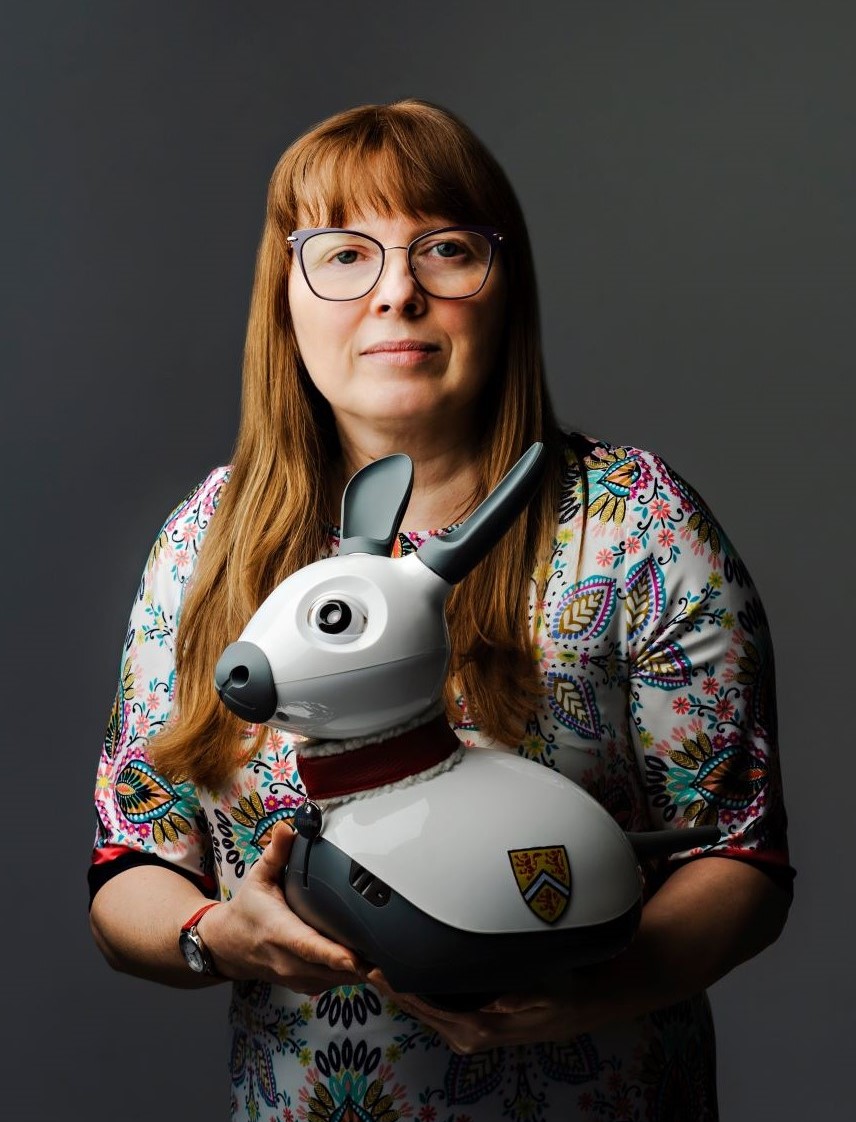}}]{Kerstin Dautenhahn} is Canada 150 Research Chair in Intelligent Robotics, Faculty of Engineering, University of Waterloo, Canada where she directs the Social and Intelligent Robotics Research Laboratory (SIRRL). She became IEEE Fellow for her contributions to Social Robotics and Human-Robot Interaction. From 2000-2018 she coordinated the Adaptive Systems Research Group at University of Hertfordshire, UK. Her main research areas are human-robot interaction, social robotics, cognitive and developmental robotics, and assistive technology, with applications of social robots as tools in education, therapy, healthcare and wellbeing.
\end{IEEEbiography}

\begin{IEEEbiography}[{\includegraphics[width=1in,height=1.25in,clip,keepaspectratio]{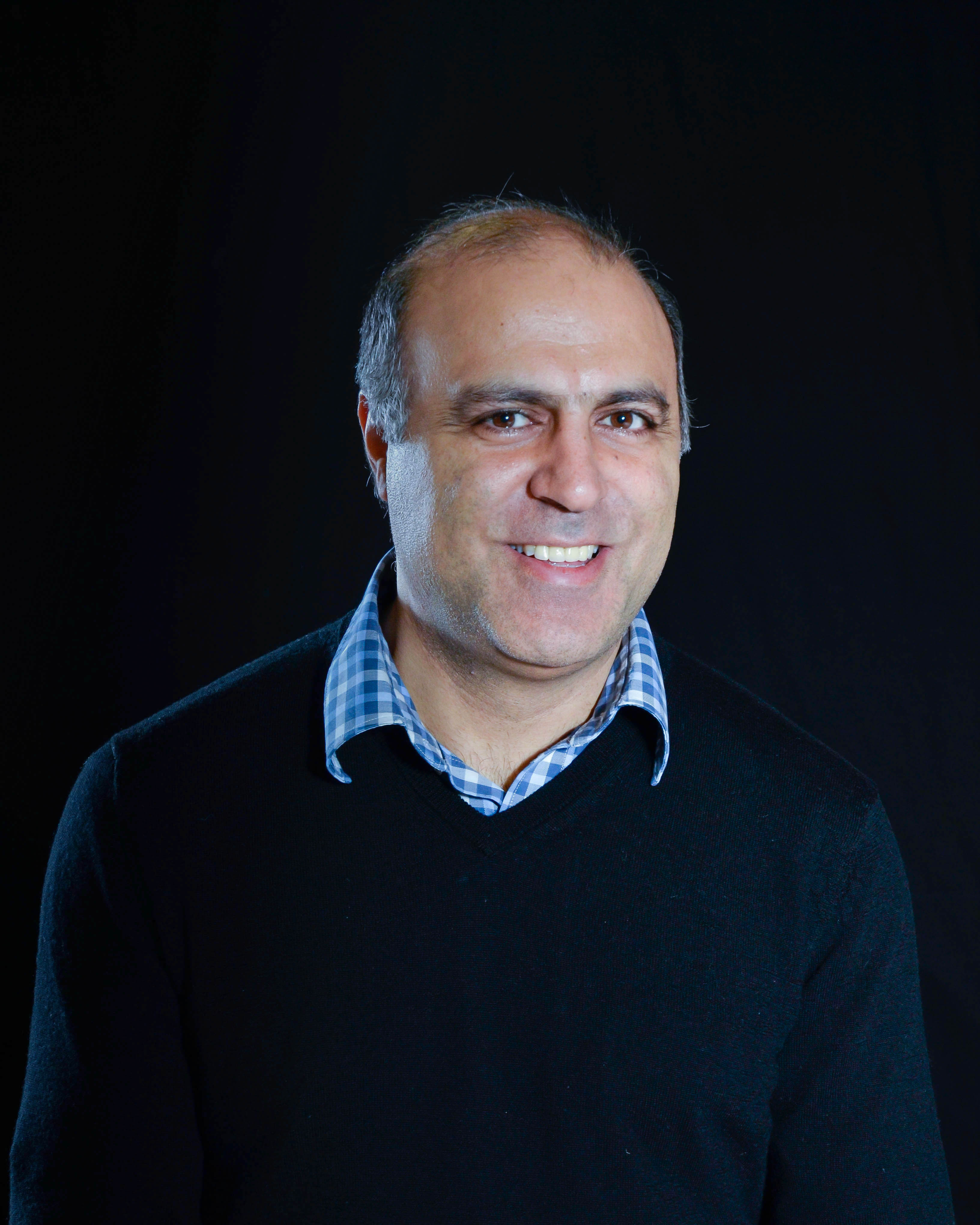}}]{Nasser L. Azad} is an Associate Professor in the Department of Systems Design Engineering at the University of Waterloo and the Director of the Automation and Intelligent Systems (AIS) Group. Before, he was a Postdoctoral Fellow at the University of California, Berkeley. Dr. Azad’s primary research interests lie in (i) intelligent, safe, and secure controls \& automation with applications to automotive systems and autonomous systems like automated vehicles and drones and (ii) innovative applications of AI methods to solve complex modeling, optimization, control, and automation problems.
\end{IEEEbiography}

\vfill

\end{document}